\documentclass[10pt,twocolumn]{article}

\usepackage{cvpr}
\usepackage{times}
\usepackage{epsfig}
\usepackage{graphicx}
\usepackage{amsmath}
\usepackage{amssymb}
\usepackage{subcaption}
\usepackage{multirow}

\usepackage{comment}
\usepackage[table]{xcolor}
\definecolor{col1}{RGB}{232, 161, 148}
\definecolor{col2}{RGB}{148, 187, 232}
\definecolor{lightblue}{RGB}{225, 225, 255}
\definecolor{darkgreen}{RGB}{0, 150, 0}
\definecolor{darkred}{RGB}{200, 0, 0}

\definecolor{dsectioncolor}{RGB}{245, 245, 245}
\definecolor{ssectioncolor}{RGB}{175, 238, 255}%
\definecolor{hrsectioncolor}{RGB}{255, 228, 196}
\definecolor{cropsectioncolor}{RGB}{255, 155, 155}
\definecolor{mssectioncolor}{RGB}{255, 200, 233}

\usepackage[pagebackref=true,breaklinks=true,letterpaper=true,colorlinks,bookmarks=false]{hyperref}

\cvprfinalcopy %

\def\cvprPaperID{****} %

\ifcvprfinal\fi
\begin{document}

\title{Image Stylization for Robust Features}

\author{Iaroslav Melekhov$^{1^*}$
\hspace{15pt}
Gabriel J. Brostow$^2$
\hspace{15pt}
Juho Kannala$^1$
\hspace{15pt}
Daniyar Turmukhambetov$^2$
\\
\vspace{3pt}
$^1$Aalto University
\hspace{20pt}
$^2$Niantic
}

\maketitle
\footnotetext[1]{
{\tt\small first.last@aalto.fi}}

\begin{abstract}
Local features that are robust to both viewpoint and appearance changes are crucial for many computer vision tasks. In this work we investigate if photorealistic image stylization improves robustness of local features to not only day-night, but also weather and season variations. We show that image stylization in addition to color augmentation is a powerful method of learning robust features. We evaluate learned features on visual localization benchmarks, outperforming state of the art baseline models despite training without ground-truth 3D correspondences using synthetic homographies only.

We use trained feature networks to compete in Long-Term Visual Localization and Map-based Localization for Autonomous Driving challenges achieving competitive scores.
\end{abstract}

\section{Introduction}~\label{sec:intro}
Detecting and describing local features (\ie interest points) for the purpose of finding correspondences between images is an important task in computer vision. The estimation of correspondences between a pair of images, \ie image matching or feature matching, is a crucial step in many pipelines in SLAM, Visual Odometry, Structure-from-Motion, Image-based localization, Panorama stitching \etc.
These computer vision tasks are the main building blocks for many applications in Augmented and Mixed Reality, Robotics and Autonomous Driving.

Hand-crafted features such as SIFT~\cite{sift} and ORB~\cite{orb} are still dominant for estimating correspondences in existing systems such as COLMAP~\cite{colmap1} and ORB-SLAM~\cite{orbslam}. Motivated by ability of neural networks to learn representations that are invariant, much focus recently has been spent on using deep learning to train neural networks to describe and/or detect local features (see a survey by Csurka and Humenberger~\cite{csurka2018handcrafted}). For local features in particular, it is desirable to have representations that are invariant or robust to nuisance changes, which are mainly due to either viewpoint variation or appearance variation due to illumination and weather changes.

Many SOTA detectors and descriptors (\eg R2D2~\cite{r2d2,r2d2_neurips}, D2-Net~\cite{d2net}, ASLfeat~\cite{aslfeat}, ContextDesc~\cite{contextdesc}, \etc) exploit large structure-from-motion reconstructions with semi-dense multi-view stereo depth estimates during their training to learn descriptors (and sometimes detectors) that are robust to viewpoint changes. These SfM reconstructions are now easier to obtain due to software packages like COLMAP~\cite{colmap1,colmap2}, but careful filtering of pixels is still required to obtain high quality depth maps~\cite{megadepth}.
In addition to SfM reconstructions, there is another computationally cheap source of training data: synthetically warped images with known homography transformations. The synthetically warped images can be generated from millions of images available from the internet~\cite{imagenet,mscoco}. Recently proposed models such as Superpoint~\cite{superpoint}, PN~\cite{yan2019unsupervised}, and R2D2~\cite{r2d2} use this source of data for training, sometimes in addition to other sources of data that capture viewpoint changes.

Learning features that are invariant to illumination or weather changes is challenging as suitable data is not easy to acquire. We show that color augmentations at training time increases robustness to appearance changes, but it is not sufficient by itself.

SfM systems can build 3D models from internet images that are captured at different times and weather conditions. Hence training on this data introduces some robustness to appearance changes. However, building 3D models from images captured under extreme changes is much more challenging, \eg reconstructing 3D models from day and night images~\cite{radenovic2016dusk}.

Another data source for appearance robustness is webcam footage archives~\cite{amos1,amos2,amospatches,verdie2015tilde}, where stationary cameras throughout the world broadcast imagery for many years. However, this source of data has its own challenges:
(i) There are moving objects in the frames, which are difficult to detect, \eg is an appearance change between two frames due to a shadow cast by a tree, or a grey car that is parked on the street?;
(ii) Very few cameras are completely stationary, as the camera mount can sometimes be readjusted or wind could introduce small, but noticeable motion;
(iii) Majority of webcams have low resolution, limited dynamic range, bad focus, etc.

Amos Patches~\cite{amospatches} proposes a pipeline to deal with these challenges with an additional manual verification at the end of the process. Despite large amount of webcam data, the resulting Amos Patches dataset consists of only 1350 images (50 images per 27 cameras).

Robustness to day-night variation using image stylization was proposed in R2D2 method~\cite{r2d2}. Day images from Aachen dataset~\cite{aachen1,sattler2018benchmarking} were converted to night images using image stylization~\cite{Li_2018_ECCV}. This resulted in improvement of descriptors for day to night feature matching, as during training each day image would have a corresponding night images with exactly the same content. Ablation study showed that this improved performance on Aachen Day-Night Localization challenge. However, only Aachen images were stylized, and only to a night style. So, appearance variations associated with weather or season changes were not explicitly modelled. Second, image stylization was not applied to web images that were used for synthetic homography training.

\begin{figure*}[ht!]
 	\centering
 	\begin{subfigure}[t]{.14\textwidth}
 		\centering
 		\includegraphics[width=\textwidth]{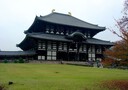}
 		\caption{Original}
 	\end{subfigure}%
 	~
 	\begin{subfigure}[t]{.14\textwidth}
 		\centering
 		\includegraphics[width=\textwidth]{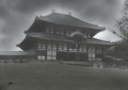}
 		\caption{cloud}
 	\end{subfigure}%
 	~
 	\begin{subfigure}[t]{.14\textwidth}
 		\centering
 		\includegraphics[width=\textwidth]{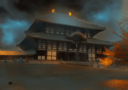}
 		\caption{dusk}
 	\end{subfigure}%
 	~
 	\begin{subfigure}[t]{.14\textwidth}
 		\centering
 		\includegraphics[width=\textwidth]{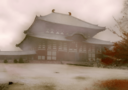}
 		\caption{mist}
 	\end{subfigure}%
 	~
 	\begin{subfigure}[t]{.14\textwidth}
 		\centering
 		\includegraphics[width=\textwidth]{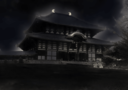}
 		\caption{night}
 	\end{subfigure}%
 	~
 	\begin{subfigure}[t]{.14\textwidth}
 		\centering
 		\includegraphics[width=\textwidth]{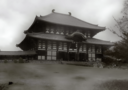}
 		\caption{rain}
 	\end{subfigure}%
 	~
 	\begin{subfigure}[t]{.14\textwidth}
 		\centering
 		\includegraphics[width=\textwidth]{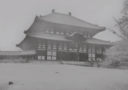}
 		\caption{snow}
 	\end{subfigure}
\caption{Example image stylizations applied to an image (a). Stylized images have extreme appearance changes, while the content of the original image is preserved. For example, the tree on the right and the details of the roof are reconizable in stylized images.}\label{fig:style:example}
\end{figure*}

We investigate how local feature robustness to appearance variation can be improved with image stylization. Image stylization allows us to generate \emph{multiple} stylizations for the same image, \eg single day image can be stylized using 10 different night images as target styles, resulting in 10 night images with the same content and viewpoint. Furthermore, the same image can be stylized to different conditions, like snow, rain or night, see Figure~\ref{fig:style:example}.

We conduct controlled experiments to compare how much gain does image stylization provide over other augmentation strategies in the literature under same evaluation conditions. We show that a model trained with synthetic homographies and image stylizations outperforms SOTA models trained with 3D correspondences. %

Finally, we demonstrate competitive scores on multiple benchmarks and challenges that evaluate visual localization under illumination, weather, and season changes.

\begin{figure}[t]
  \centering
    \includegraphics[width=.9\linewidth]{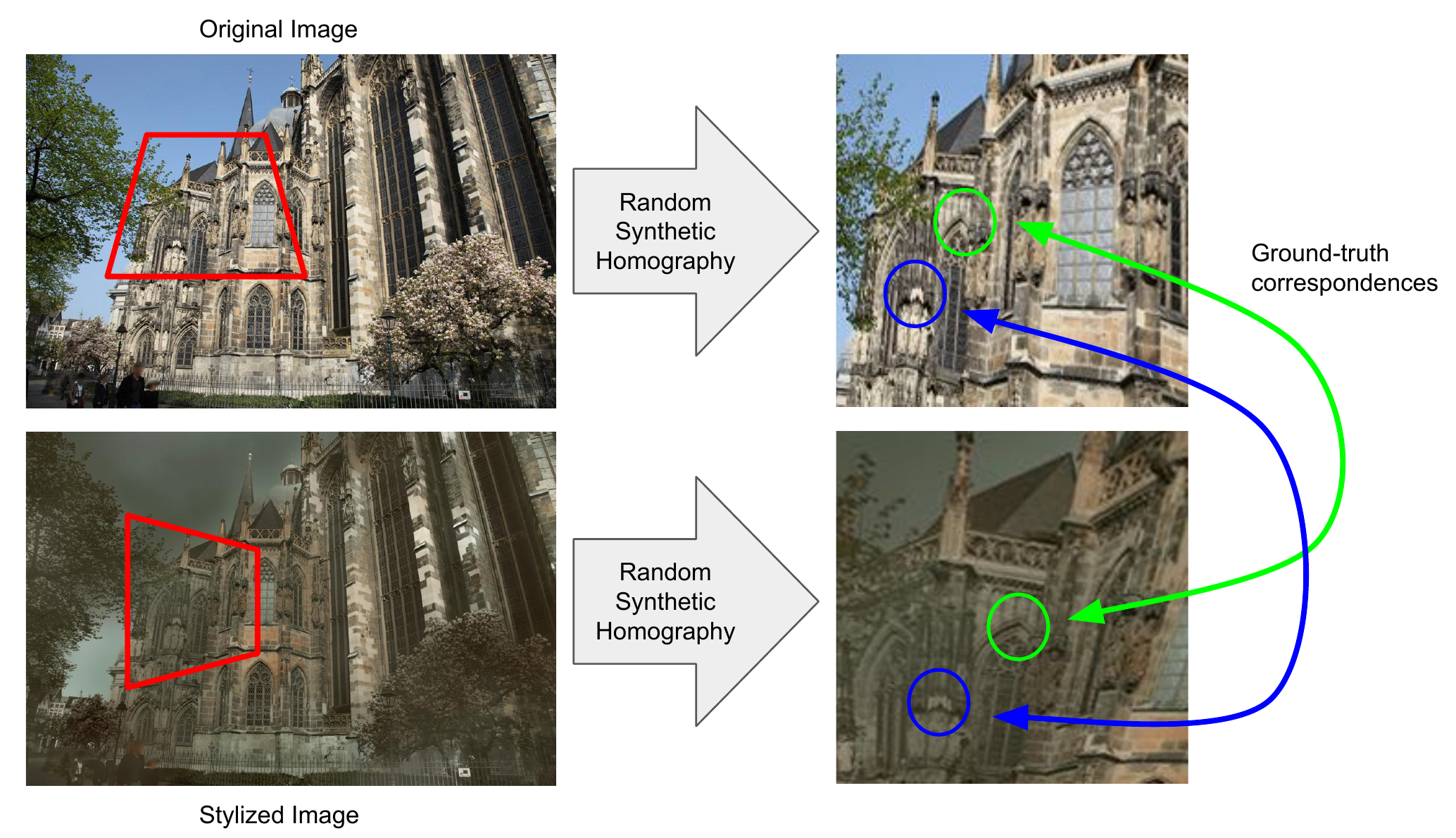}
 \caption{Method overview. We train feature networks using original and stylized image pairs. Crops are extracted from the original and stylized images using random homographies. Image stylization and synthetic homography augmentation together allow us to learn features that are robust to appearance and viewpoint changes. }
 \label{fig:f_pipeline}
\end{figure}

\section{Method}\label{sec:method}
For increased robustness of local features we train feature networks using original-stylized image pairs. Here, we describe how we generate stylized images and losses used for feature network training.

\subsection{Image Stylization}

Style transfer method takes as input 2 images, $C$ and $S$ and produces an image $C_S$ which has content of image $C$, but the style of image $S$.
Similar to R2D2~\cite{r2d2} we use \cite{Li_2018_ECCV} for image stylization, as it preserves edges and objects after stylization, unlike other methods designed for artistic style transfer, \eg~\cite{gatys2015neural}.

\begin{figure*}[ht!]
\centering
\begin{minipage}{.13\textwidth}
    \centering
    \includegraphics[height=0.06\textheight]{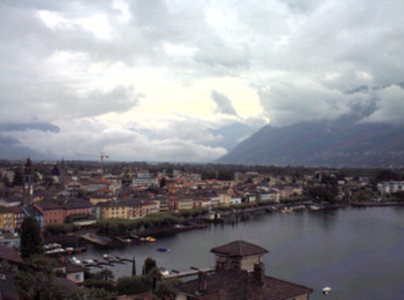}
\end{minipage}
\begin{minipage}{.13\textwidth}
    \centering
    \includegraphics[height=0.06\textheight]{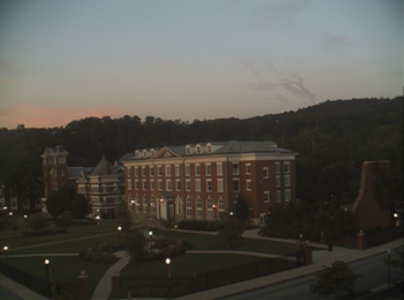}
\end{minipage}
\begin{minipage}{.13\textwidth}
    \centering
    \includegraphics[height=0.06\textheight]{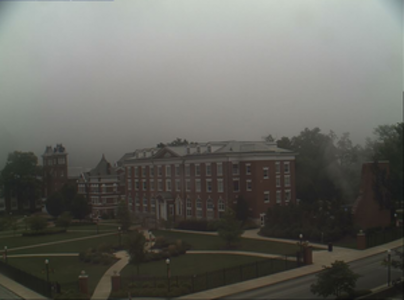}
\end{minipage}
\begin{minipage}{.13\textwidth}
    \centering
    \includegraphics[height=0.06\textheight]{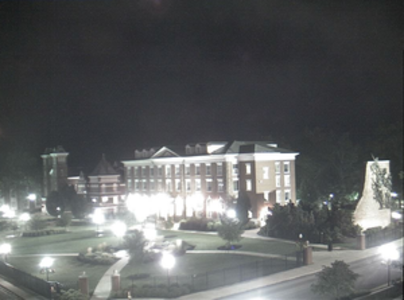}
\end{minipage}
\begin{minipage}{.13\textwidth}
    \centering
    \includegraphics[height=0.06\textheight]{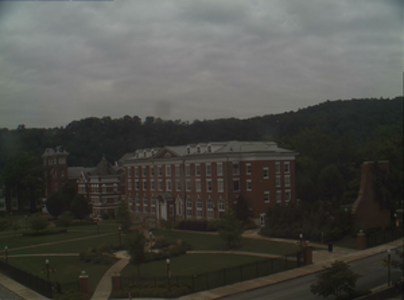}
\end{minipage}
\begin{minipage}{.13\textwidth}
    \centering
    \includegraphics[height=0.06\textheight]{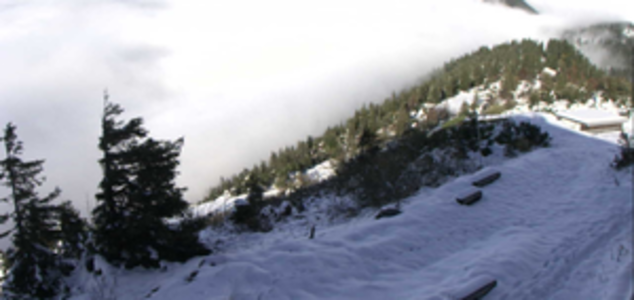}
\end{minipage}

\centering
\begin{minipage}{.13\textwidth}
    \centering
    \includegraphics[height=0.06\textheight]{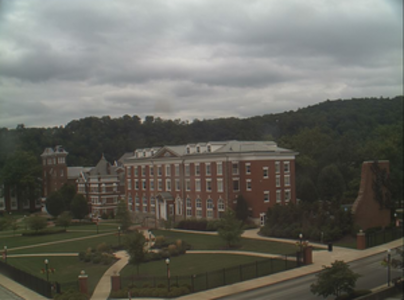}
\end{minipage}
\begin{minipage}{.13\textwidth}
    \centering
    \includegraphics[height=0.06\textheight]{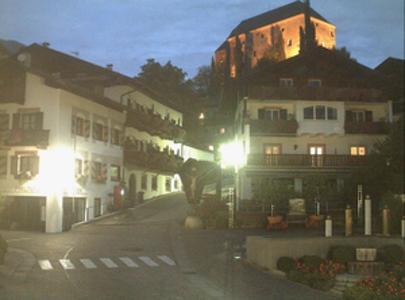}
\end{minipage}
\begin{minipage}{.13\textwidth}
    \centering
    \includegraphics[height=0.06\textheight]{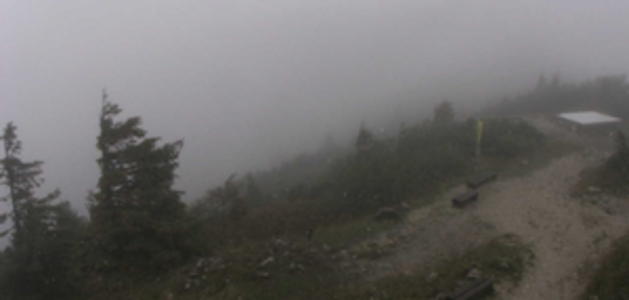}
\end{minipage}
\begin{minipage}{.13\textwidth}
    \centering
    \includegraphics[height=0.06\textheight]{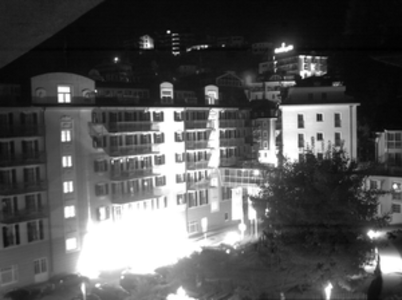}
\end{minipage}
\begin{minipage}{.13\textwidth}
    \centering
    \includegraphics[height=0.06\textheight]{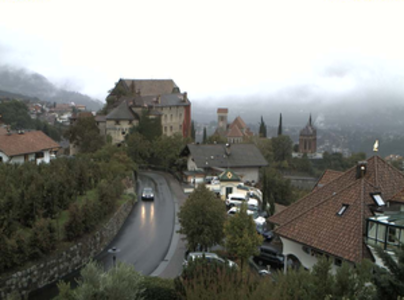}
\end{minipage}
\begin{minipage}{.13\textwidth}
    \centering
    \includegraphics[height=0.06\textheight]{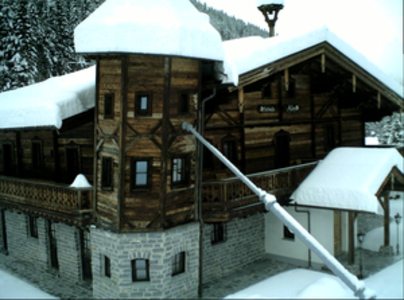}
\end{minipage}

\centering
\begin{minipage}{.13\textwidth}
    \centering
    \includegraphics[height=0.06\textheight]{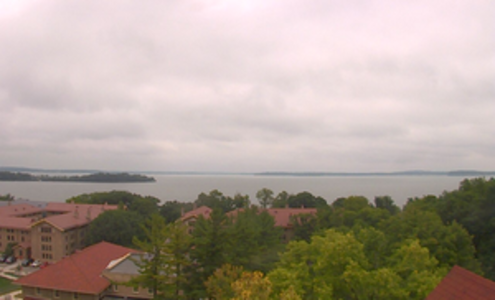}
\end{minipage}
\begin{minipage}{.13\textwidth}
    \centering
    \includegraphics[height=0.06\textheight]{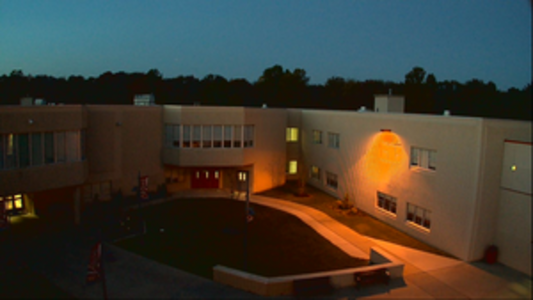}
\end{minipage}
\begin{minipage}{.13\textwidth}
    \centering
    \includegraphics[height=0.06\textheight]{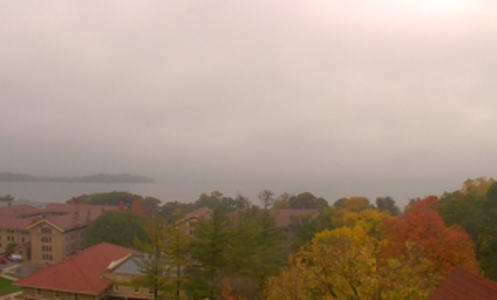}
\end{minipage}
\begin{minipage}{.13\textwidth}
    \centering
    \includegraphics[height=0.06\textheight]{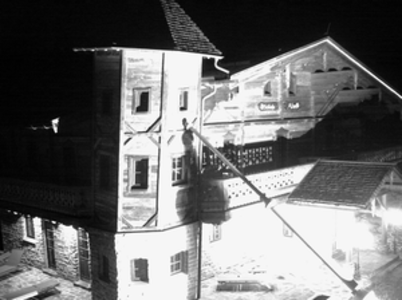}
\end{minipage}
\begin{minipage}{.13\textwidth}
    \centering
    \includegraphics[height=0.06\textheight]{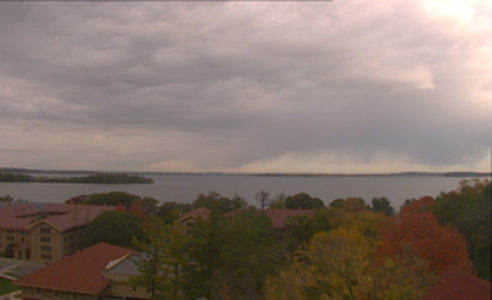}
\end{minipage}
\begin{minipage}{.13\textwidth}
    \centering
    \includegraphics[height=0.06\textheight]{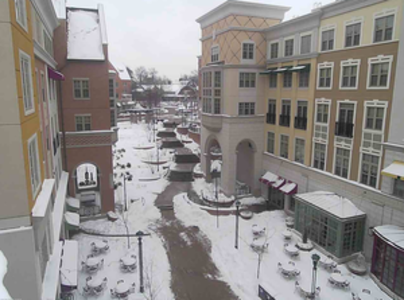}
\end{minipage}

\centering
\begin{minipage}{.13\textwidth}
    \centering
    \includegraphics[height=0.06\textheight]{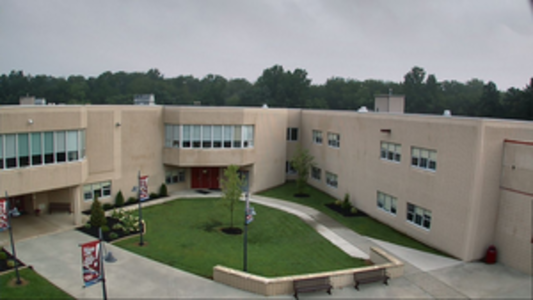}
\end{minipage}
\begin{minipage}{.13\textwidth}
    \centering
    \includegraphics[height=0.06\textheight]{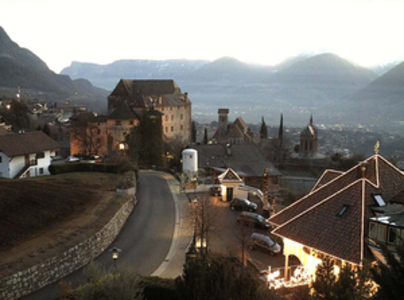}
\end{minipage}
\begin{minipage}{.13\textwidth}
    \centering
    \includegraphics[height=0.06\textheight]{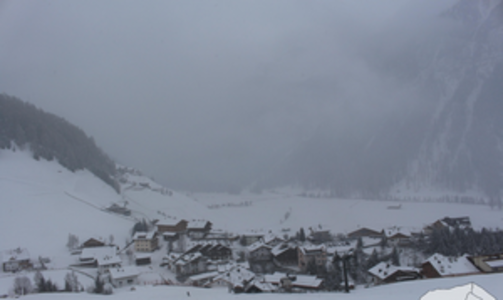}
\end{minipage}
\begin{minipage}{.13\textwidth}
    \centering
    \includegraphics[height=0.06\textheight]{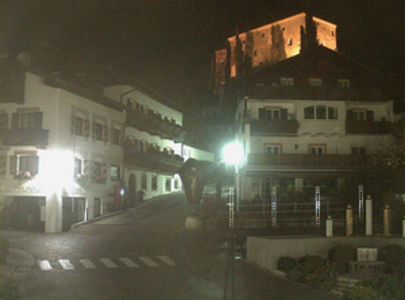}
\end{minipage}
\begin{minipage}{.13\textwidth}
    \centering
    \includegraphics[height=0.06\textheight]{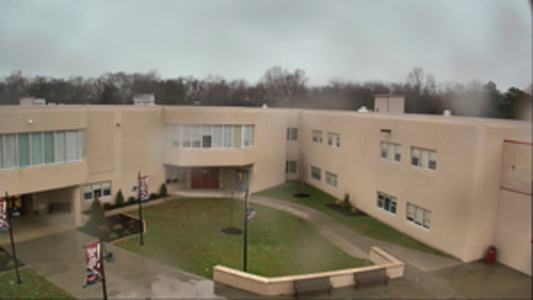}
\end{minipage}
\begin{minipage}{.13\textwidth}
    \centering
    \includegraphics[height=0.06\textheight]{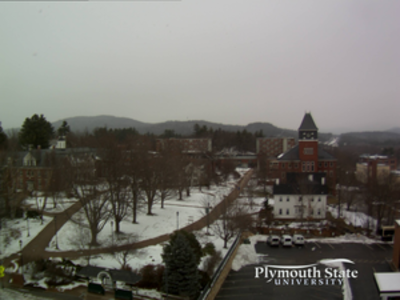}
\end{minipage}

\centering
\begin{minipage}{.13\textwidth}
    \centering
    \includegraphics[height=0.06\textheight]{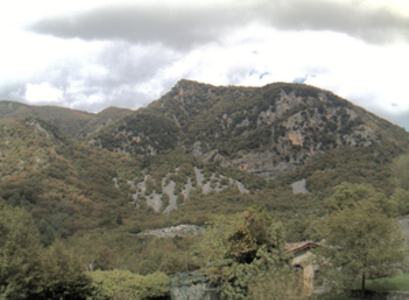}
\end{minipage}
\begin{minipage}{.13\textwidth}
    \centering
    \includegraphics[height=0.06\textheight]{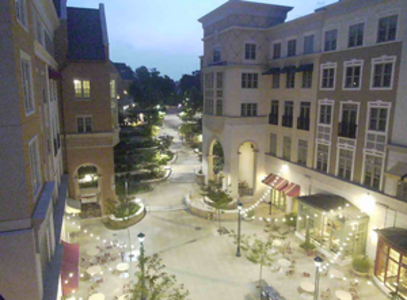}
\end{minipage}
\begin{minipage}{.13\textwidth}
    \centering
    \includegraphics[height=0.06\textheight]{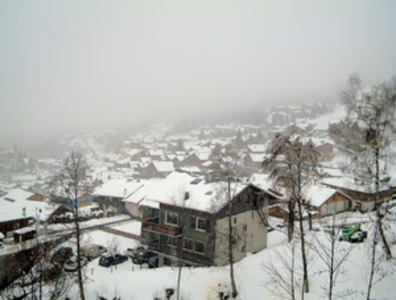}
\end{minipage}
\begin{minipage}{.13\textwidth}
    \centering
    \includegraphics[height=0.06\textheight]{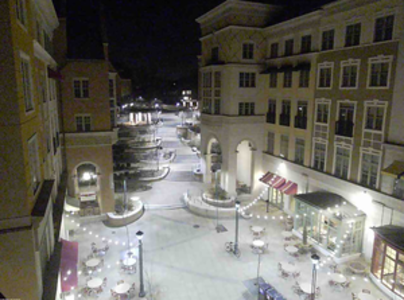}
\end{minipage}
\begin{minipage}{.13\textwidth}
    \centering
    \includegraphics[height=0.06\textheight]{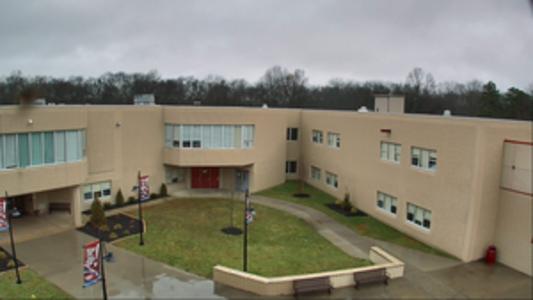}
\end{minipage}
\begin{minipage}{.13\textwidth}
    \centering
    \includegraphics[height=0.06\textheight]{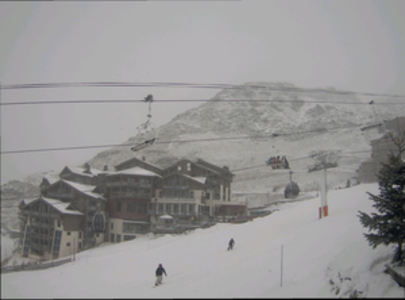}
\end{minipage}

\centering
\begin{minipage}{.13\textwidth}
    \centering
    \includegraphics[height=0.06\textheight]{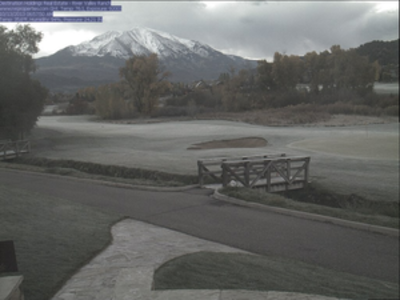}
\end{minipage}
\begin{minipage}{.13\textwidth}
    \centering
    \includegraphics[height=0.06\textheight]{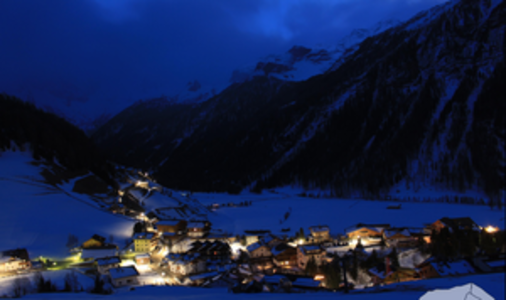}
\end{minipage}
\begin{minipage}{.13\textwidth}
    \centering
    \includegraphics[height=0.06\textheight]{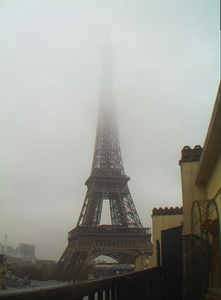}
\end{minipage}
\begin{minipage}{.13\textwidth}
    \centering
    \includegraphics[height=0.06\textheight]{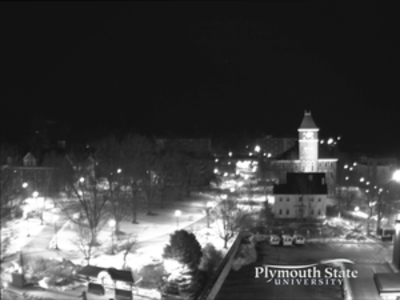}
\end{minipage}
\begin{minipage}{.13\textwidth}
    \centering
    \includegraphics[height=0.06\textheight]{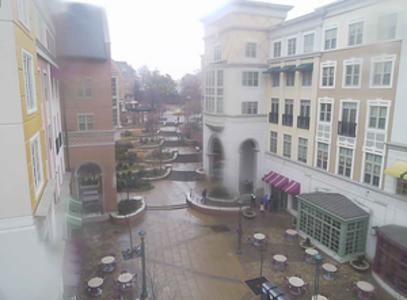}
\end{minipage}
\begin{minipage}{.13\textwidth}
    \centering
    \includegraphics[height=0.06\textheight]{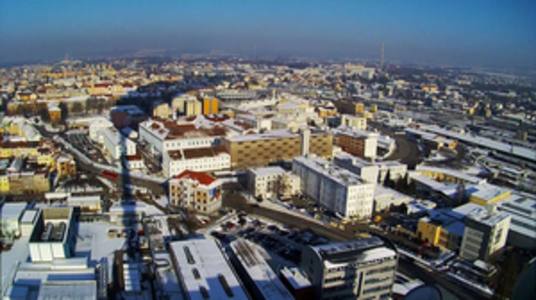}
\end{minipage}

\centering
\begin{minipage}{.13\textwidth}
    \centering
    \includegraphics[height=0.06\textheight]{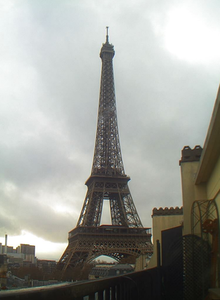}
\end{minipage}
\begin{minipage}{.13\textwidth}
    \centering
    \includegraphics[height=0.06\textheight]{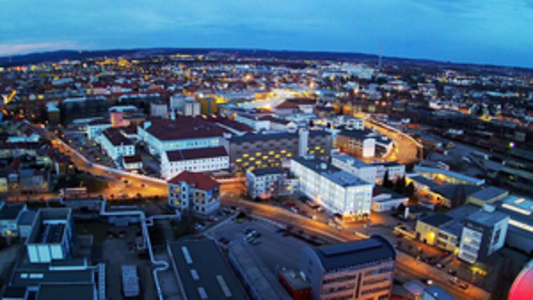}
\end{minipage}
\begin{minipage}{.13\textwidth}
    \centering
    \includegraphics[height=0.06\textheight]{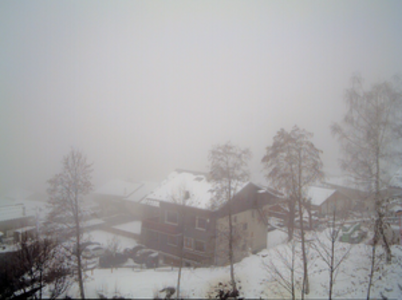}
\end{minipage}
\begin{minipage}{.13\textwidth}
    \centering
    \includegraphics[height=0.06\textheight]{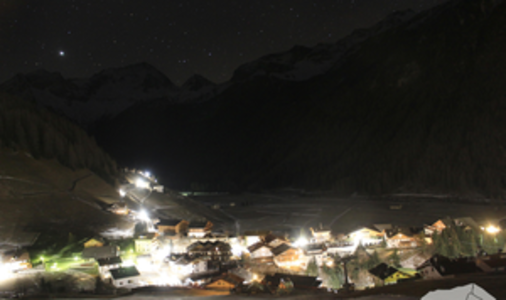}
\end{minipage}
\begin{minipage}{.13\textwidth}
    \centering
    \includegraphics[height=0.06\textheight]{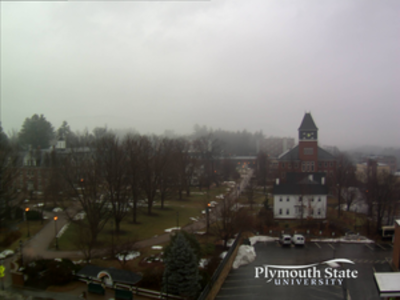}
\end{minipage}
\begin{minipage}{.13\textwidth}
    \centering
    \includegraphics[height=0.06\textheight]{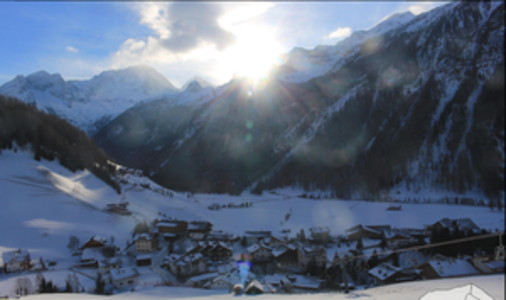}
\end{minipage}

\centering
\begin{minipage}{.13\textwidth}
    \centering
    \includegraphics[height=0.06\textheight]{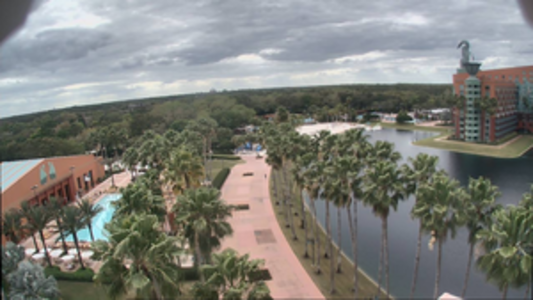}
\end{minipage}
\begin{minipage}{.13\textwidth}
    \centering
    \includegraphics[height=0.06\textheight]{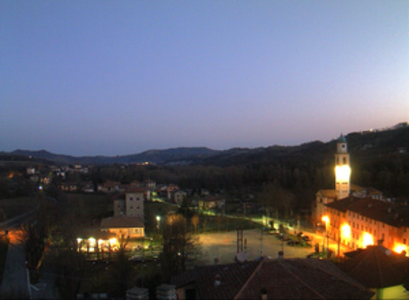}
\end{minipage}
\begin{minipage}{.13\textwidth}
    \centering
    \includegraphics[height=0.06\textheight]{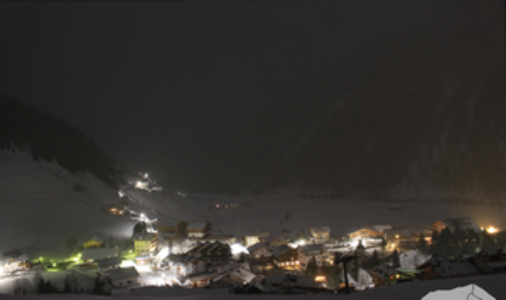}
\end{minipage}
\begin{minipage}{.13\textwidth}
    \centering
    \includegraphics[height=0.06\textheight]{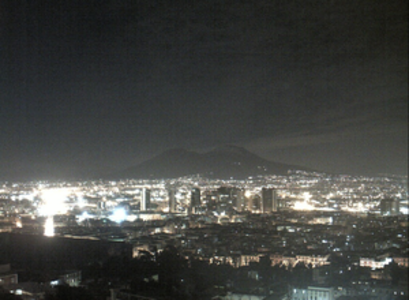}
\end{minipage}
\begin{minipage}{.13\textwidth}
    \centering
    \includegraphics[height=0.06\textheight]{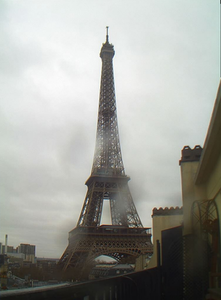}
\end{minipage}
\begin{minipage}{.13\textwidth}
    \centering
    \includegraphics[height=0.06\textheight]{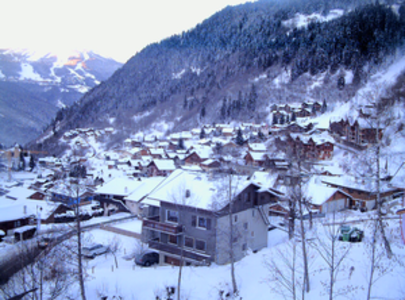}
\end{minipage}

\centering
\begin{minipage}{.13\textwidth}
    \centering
    \includegraphics[height=0.06\textheight]{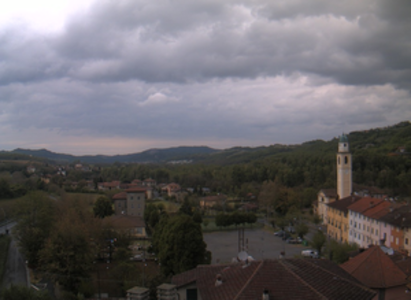}
\end{minipage}
\begin{minipage}{.13\textwidth}
    \centering
    \includegraphics[height=0.06\textheight]{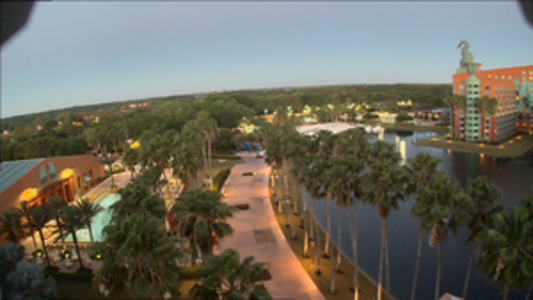}
\end{minipage}
\begin{minipage}{.13\textwidth}
    \centering
    \includegraphics[height=0.06\textheight]{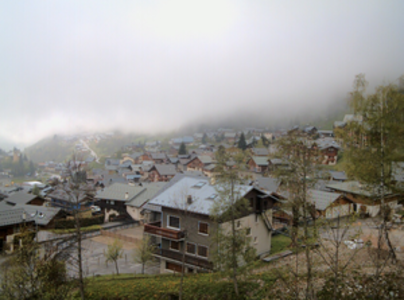}
\end{minipage}
\begin{minipage}{.13\textwidth}
    \centering
    \includegraphics[height=0.06\textheight]{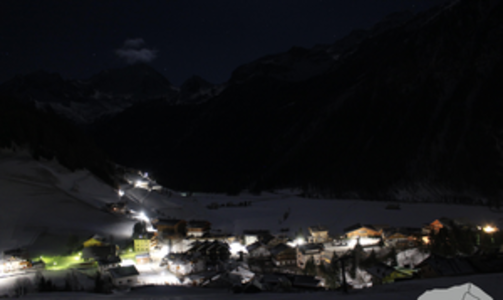}
\end{minipage}
\begin{minipage}{.13\textwidth}
    \centering
    \includegraphics[height=0.06\textheight]{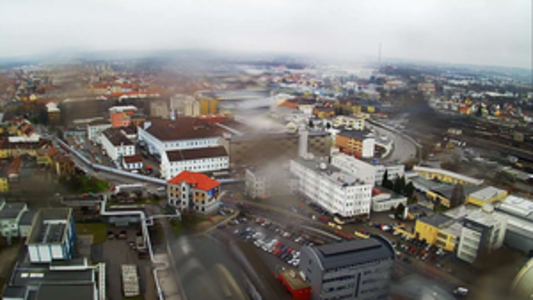}
\end{minipage}
\begin{minipage}{.13\textwidth}
    \centering
    \includegraphics[height=0.06\textheight]{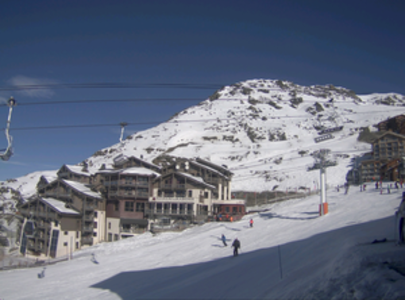}
\end{minipage}

\centering
\begin{minipage}{.13\textwidth}
    \centering
    \includegraphics[height=0.06\textheight]{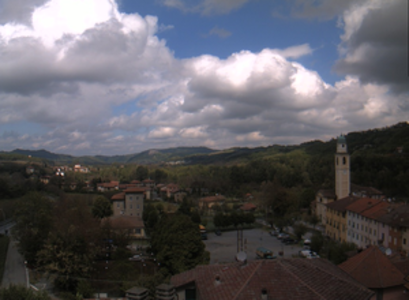}
\end{minipage}
\begin{minipage}{.13\textwidth}
    \centering
    \includegraphics[height=0.06\textheight]{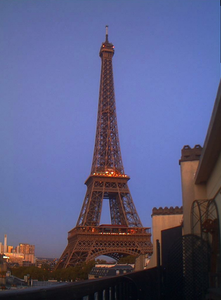}
\end{minipage}
\begin{minipage}{.13\textwidth}
    \centering
    \includegraphics[height=0.06\textheight]{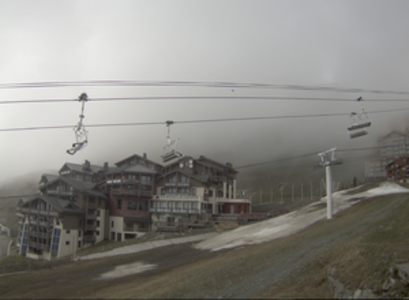}
\end{minipage}
\begin{minipage}{.13\textwidth}
    \centering
    \includegraphics[height=0.06\textheight]{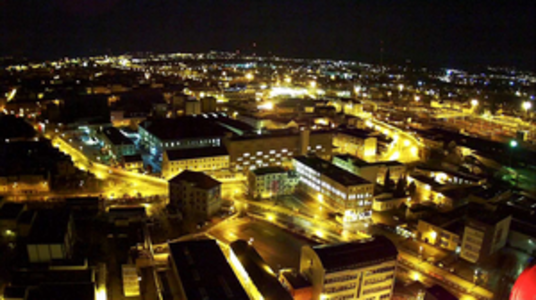}
\end{minipage}
\begin{minipage}{.13\textwidth}
    \centering
    \includegraphics[height=0.06\textheight]{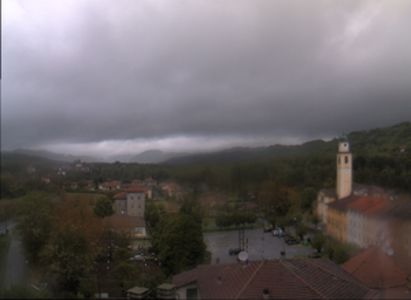}
\end{minipage}
\begin{minipage}{.13\textwidth}
    \centering
    \includegraphics[height=0.06\textheight]{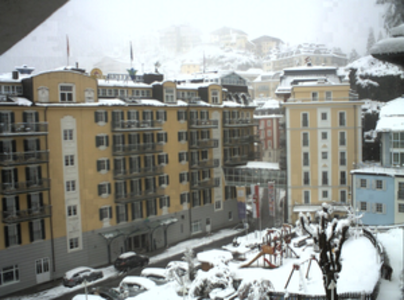}
\end{minipage}

\vspace{2pt}
\centering
\begin{minipage}{.13\textwidth}
    \centering
    cloudy
\end{minipage}
\begin{minipage}{.13\textwidth}
    \centering
    dusk
\end{minipage}
\begin{minipage}{.13\textwidth}
    \centering
    mist
\end{minipage}
\begin{minipage}{.13\textwidth}
    \centering
    night
\end{minipage}
\begin{minipage}{.13\textwidth}
    \centering
    rainy
\end{minipage}
\begin{minipage}{.13\textwidth}
    \centering
    snow
\end{minipage}

\caption{Style images manually selected from contributing views of Amos Patches dataset~\cite{amospatches}. See Figure~\ref{fig:stylized} for corresponding stylization results. There are 10 style examples for each style category.}
\label{fig:styles}
\end{figure*}

We model appearance variation in 6 styles/categories: cloud, dusk, mist, night, rain, and snow. For each style, there are 10 style images (see Figure~\ref{fig:styles}), which were manually selected from the set of contributing views of Amos Patches dataset~\cite{amospatches}.
For each image in the training set, we generate 60 stylized images, 10 for each of 6 style categories.

\begin{figure*}
\centering
\begin{minipage}{.8\textwidth}
\textbf{(a) Original Image}\\
\end{minipage}
\\
\begin{minipage}{.8\textwidth}
    \includegraphics[height=0.06\textheight]{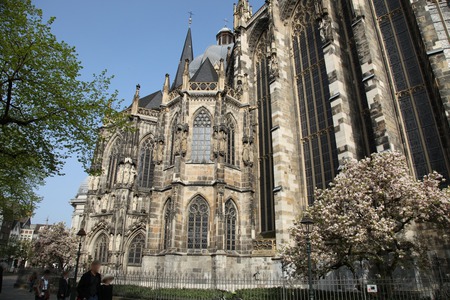}
\end{minipage}

\centering
\begin{minipage}{.8\textwidth}
    \centering
    \hrulefill\vspace{15pt}\par
\setbox0=\hbox{%
\begin{tabular}{|l|}
\\
\hline
\end{tabular}
}
\end{minipage}
\\
\vspace{-3pt}
\centering
\begin{minipage}{.8\textwidth}
\textbf{(b) R2D2 stylization results}\\
\end{minipage}
\\
\begin{minipage}{.8\textwidth}
    \includegraphics[height=0.06\textheight]{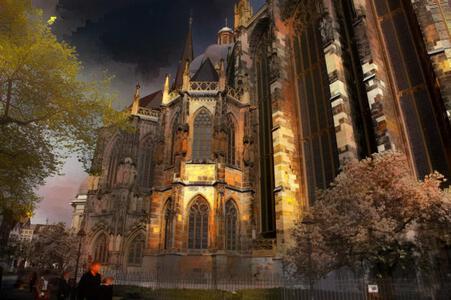}
\end{minipage}

\centering
\begin{minipage}{.8\textwidth}
    \centering
    \hrulefill\vspace{15pt}\par
\setbox0=\hbox{%
\begin{tabular}{|l|}
\\
\hline
\end{tabular}
}
\end{minipage}
\\
\vspace{-3pt}
\centering
\begin{minipage}{.8\textwidth}
\textbf{(c) Our stylization results}\\
\end{minipage}
\\

\centering
\begin{minipage}{.13\textwidth}
    \centering
    \includegraphics[height=0.06\textheight]{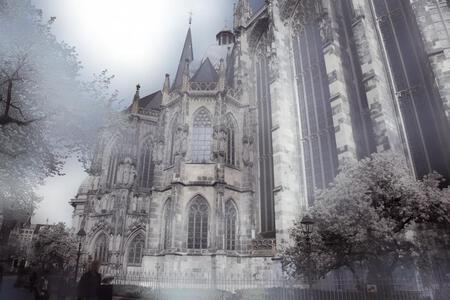}
\end{minipage}
\begin{minipage}{.13\textwidth}
    \centering
    \includegraphics[height=0.06\textheight]{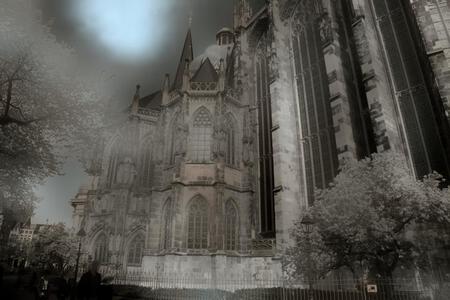}
\end{minipage}
\begin{minipage}{.13\textwidth}
    \centering
    \includegraphics[height=0.06\textheight]{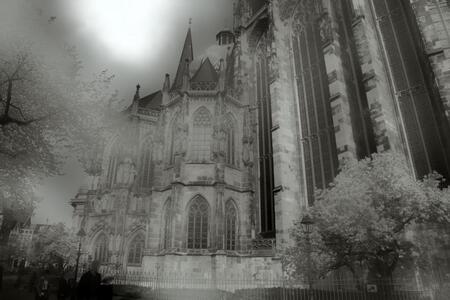}
\end{minipage}
\begin{minipage}{.13\textwidth}
    \centering
    \includegraphics[height=0.06\textheight]{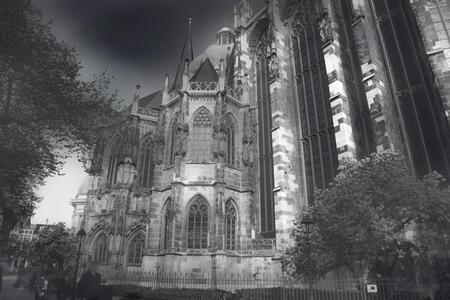}
\end{minipage}
\begin{minipage}{.13\textwidth}
    \centering
    \includegraphics[height=0.06\textheight]{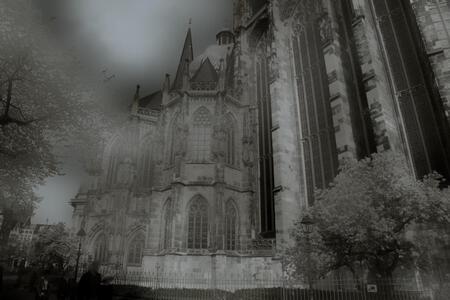}
\end{minipage}
\begin{minipage}{.13\textwidth}
    \centering
    \includegraphics[height=0.06\textheight]{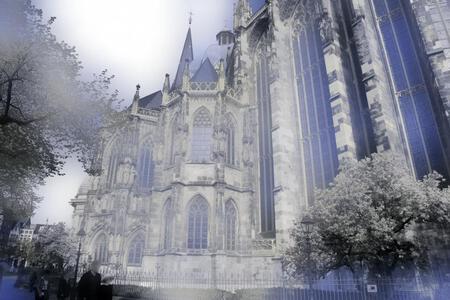}
\end{minipage}

\centering
\begin{minipage}{.13\textwidth}
    \centering
    \includegraphics[height=0.06\textheight]{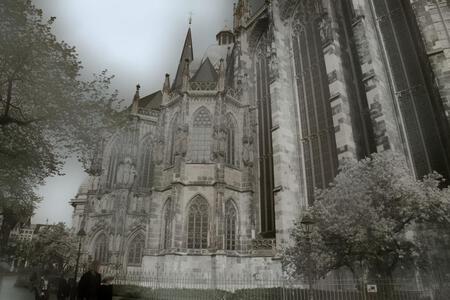}
\end{minipage}
\begin{minipage}{.13\textwidth}
    \centering
    \includegraphics[height=0.06\textheight]{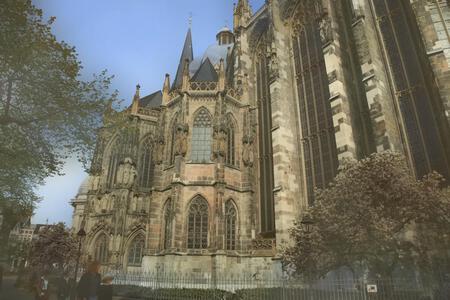}
\end{minipage}
\begin{minipage}{.13\textwidth}
    \centering
    \includegraphics[height=0.06\textheight]{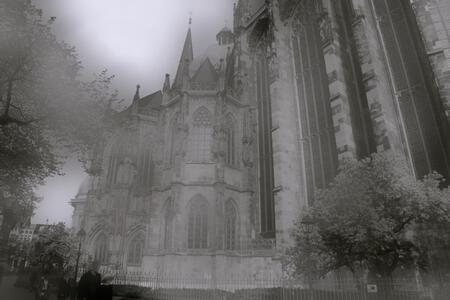}
\end{minipage}
\begin{minipage}{.13\textwidth}
    \centering
    \includegraphics[height=0.06\textheight]{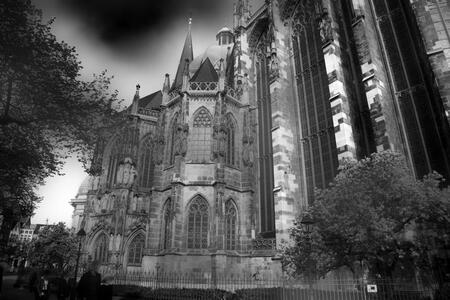}
\end{minipage}
\begin{minipage}{.13\textwidth}
    \centering
    \includegraphics[height=0.06\textheight]{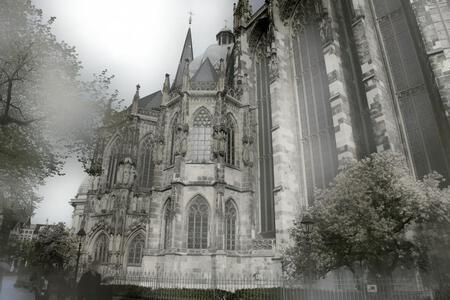}
\end{minipage}
\begin{minipage}{.13\textwidth}
    \centering
    \includegraphics[height=0.06\textheight]{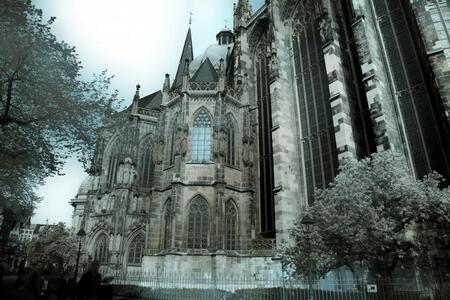}
\end{minipage}

\centering
\begin{minipage}{.13\textwidth}
    \centering
    \includegraphics[height=0.06\textheight]{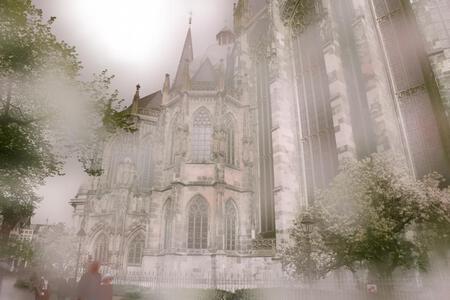}
\end{minipage}
\begin{minipage}{.13\textwidth}
    \centering
    \includegraphics[height=0.06\textheight]{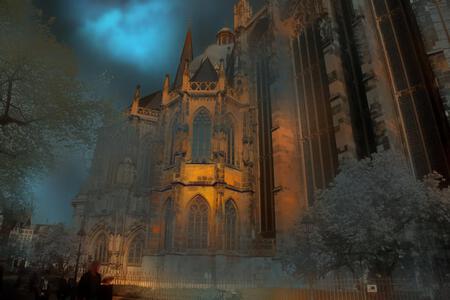}
\end{minipage}
\begin{minipage}{.13\textwidth}
    \centering
    \includegraphics[height=0.06\textheight]{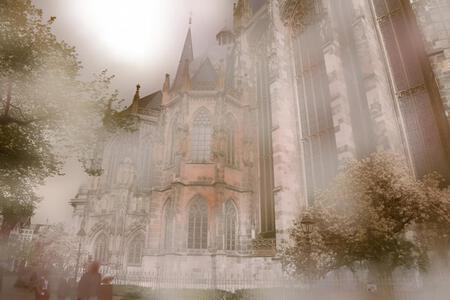}
\end{minipage}
\begin{minipage}{.13\textwidth}
    \centering
    \includegraphics[height=0.06\textheight]{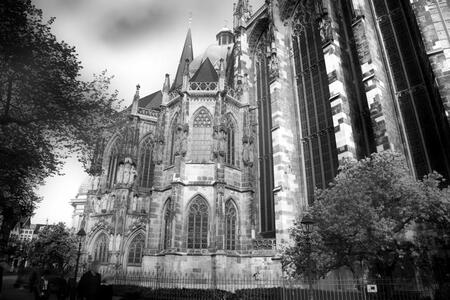}
\end{minipage}
\begin{minipage}{.13\textwidth}
    \centering
    \includegraphics[height=0.06\textheight]{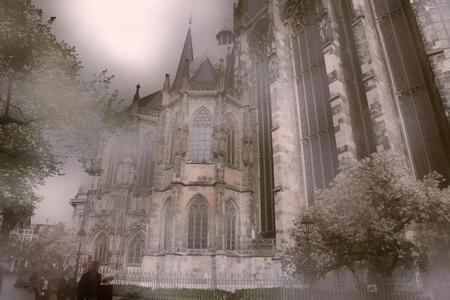}
\end{minipage}
\begin{minipage}{.13\textwidth}
    \centering
    \includegraphics[height=0.06\textheight]{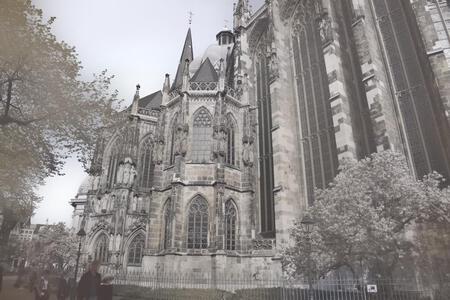}
\end{minipage}

\centering
\begin{minipage}{.13\textwidth}
    \centering
    \includegraphics[height=0.06\textheight]{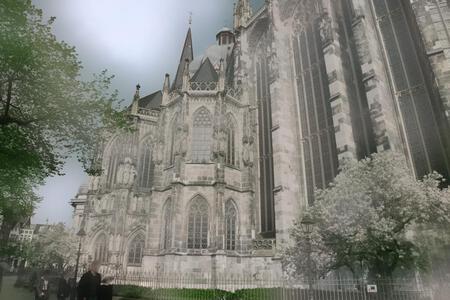}
\end{minipage}
\begin{minipage}{.13\textwidth}
    \centering
    \includegraphics[height=0.06\textheight]{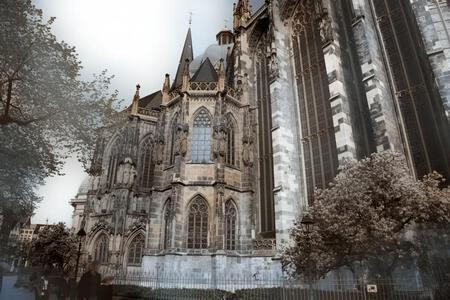}
\end{minipage}
\begin{minipage}{.13\textwidth}
    \centering
    \includegraphics[height=0.06\textheight]{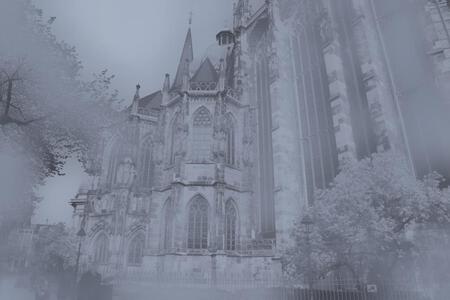}
\end{minipage}
\begin{minipage}{.13\textwidth}
    \centering
    \includegraphics[height=0.06\textheight]{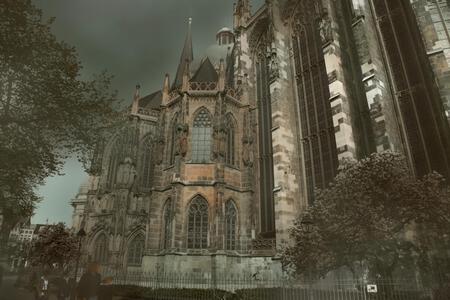}
\end{minipage}
\begin{minipage}{.13\textwidth}
    \centering
    \includegraphics[height=0.06\textheight]{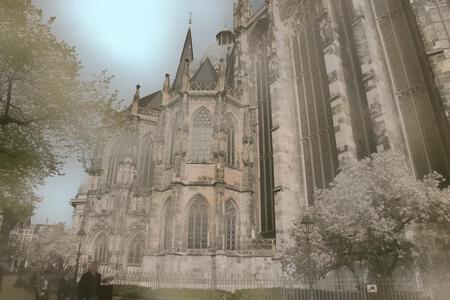}
\end{minipage}
\begin{minipage}{.13\textwidth}
    \centering
    \includegraphics[height=0.06\textheight]{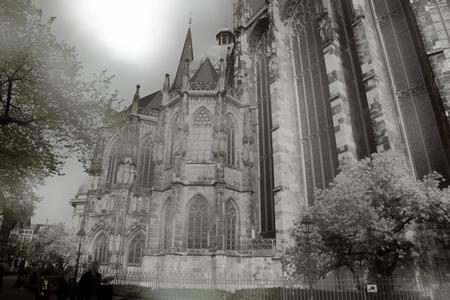}
\end{minipage}

\centering
\begin{minipage}{.13\textwidth}
    \centering
    \includegraphics[height=0.06\textheight]{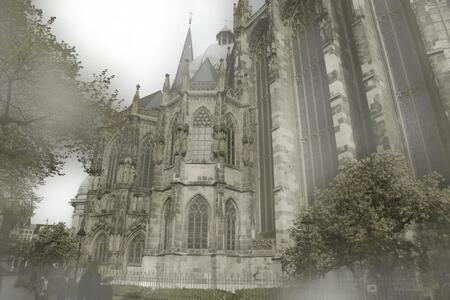}
\end{minipage}
\begin{minipage}{.13\textwidth}
    \centering
    \includegraphics[height=0.06\textheight]{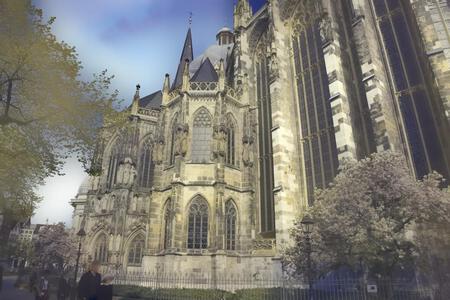}
\end{minipage}
\begin{minipage}{.13\textwidth}
    \centering
    \includegraphics[height=0.06\textheight]{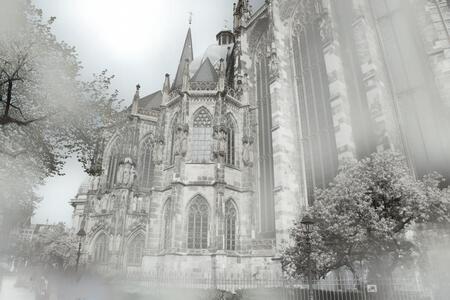}
\end{minipage}
\begin{minipage}{.13\textwidth}
    \centering
    \includegraphics[height=0.06\textheight]{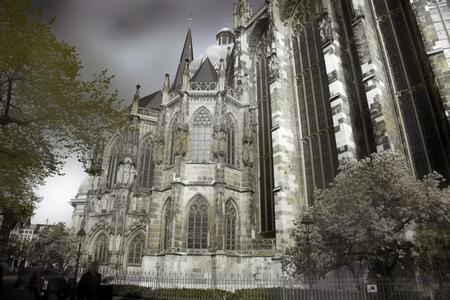}
\end{minipage}
\begin{minipage}{.13\textwidth}
    \centering
    \includegraphics[height=0.06\textheight]{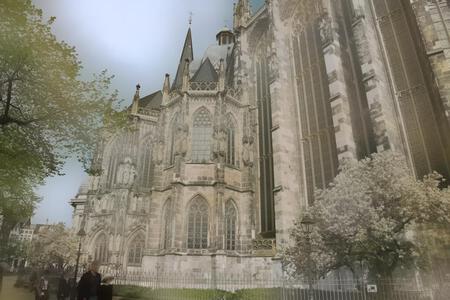}
\end{minipage}
\begin{minipage}{.13\textwidth}
    \centering
    \includegraphics[height=0.06\textheight]{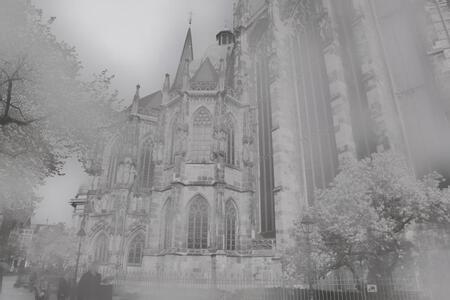}
\end{minipage}

\centering
\begin{minipage}{.13\textwidth}
    \centering
    \includegraphics[height=0.06\textheight]{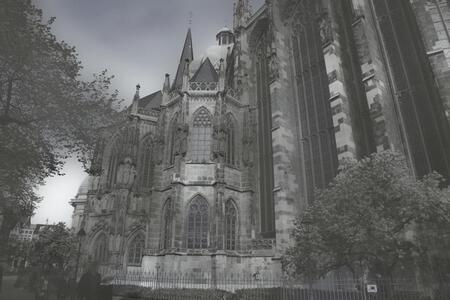}
\end{minipage}
\begin{minipage}{.13\textwidth}
    \centering
    \includegraphics[height=0.06\textheight]{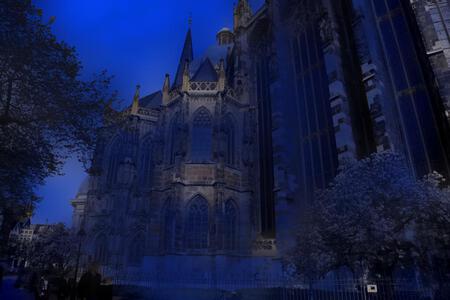}
\end{minipage}
\begin{minipage}{.13\textwidth}
    \centering
    \includegraphics[height=0.06\textheight]{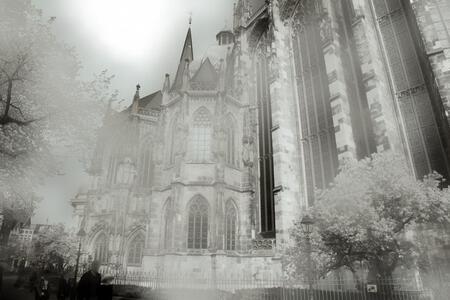}
\end{minipage}
\begin{minipage}{.13\textwidth}
    \centering
    \includegraphics[height=0.06\textheight]{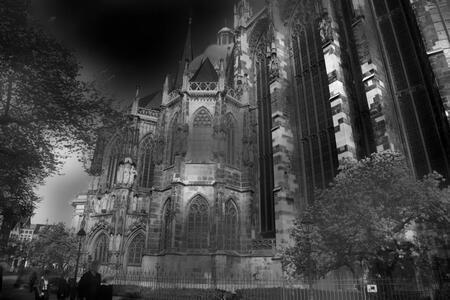}
\end{minipage}
\begin{minipage}{.13\textwidth}
    \centering
    \includegraphics[height=0.06\textheight]{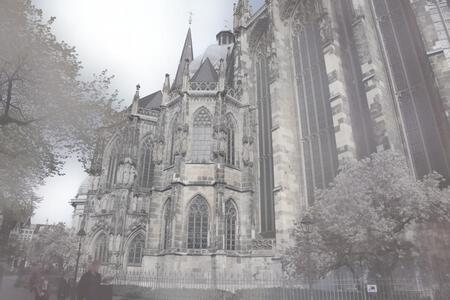}
\end{minipage}
\begin{minipage}{.13\textwidth}
    \centering
    \includegraphics[height=0.06\textheight]{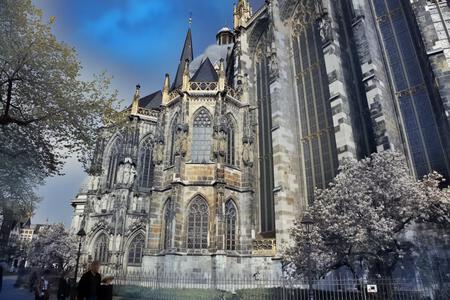}
\end{minipage}

\centering
\begin{minipage}{.13\textwidth}
    \centering
    \includegraphics[height=0.06\textheight]{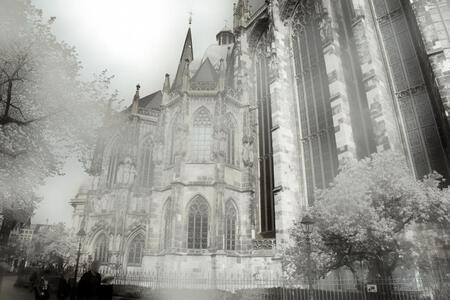}
\end{minipage}
\begin{minipage}{.13\textwidth}
    \centering
    \includegraphics[height=0.06\textheight]{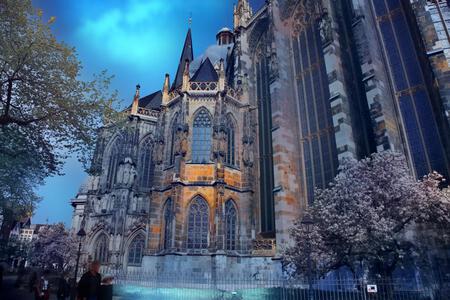}
\end{minipage}
\begin{minipage}{.13\textwidth}
    \centering
    \includegraphics[height=0.06\textheight]{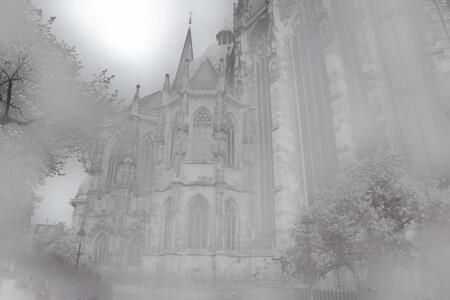}
\end{minipage}
\begin{minipage}{.13\textwidth}
    \centering
    \includegraphics[height=0.06\textheight]{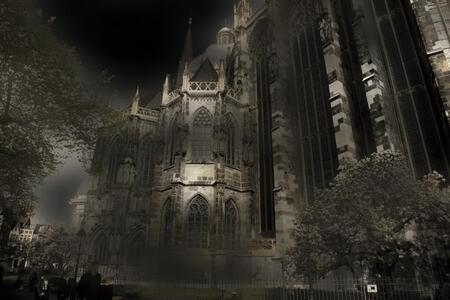}
\end{minipage}
\begin{minipage}{.13\textwidth}
    \centering
    \includegraphics[height=0.06\textheight]{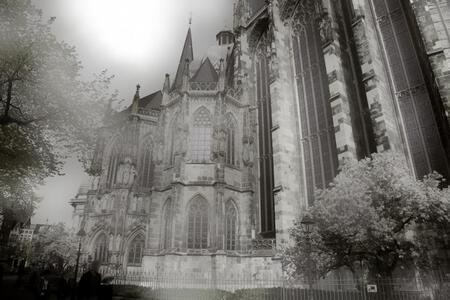}
\end{minipage}
\begin{minipage}{.13\textwidth}
    \centering
    \includegraphics[height=0.06\textheight]{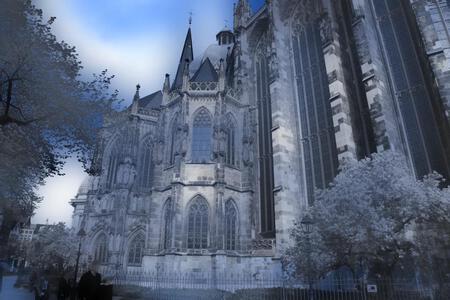}
\end{minipage}

\centering
\begin{minipage}{.13\textwidth}
    \centering
    \includegraphics[height=0.06\textheight]{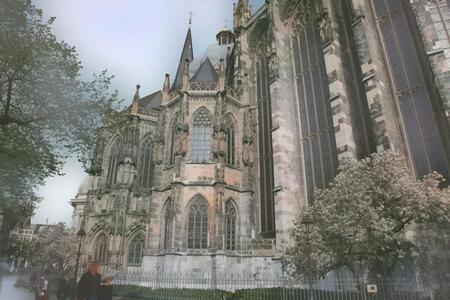}
\end{minipage}
\begin{minipage}{.13\textwidth}
    \centering
    \includegraphics[height=0.06\textheight]{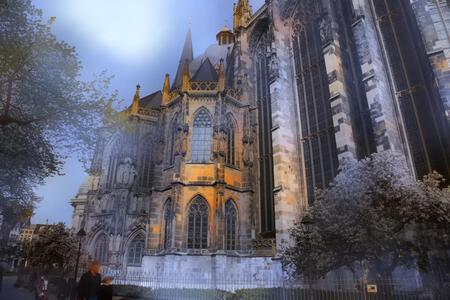}
\end{minipage}
\begin{minipage}{.13\textwidth}
    \centering
    \includegraphics[height=0.06\textheight]{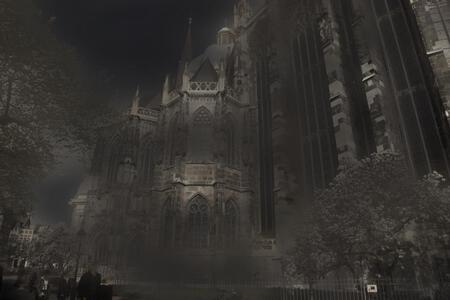}
\end{minipage}
\begin{minipage}{.13\textwidth}
    \centering
    \includegraphics[height=0.06\textheight]{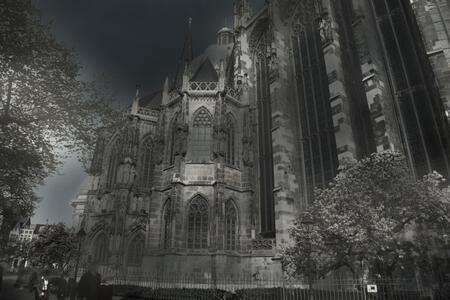}
\end{minipage}
\begin{minipage}{.13\textwidth}
    \centering
    \includegraphics[height=0.06\textheight]{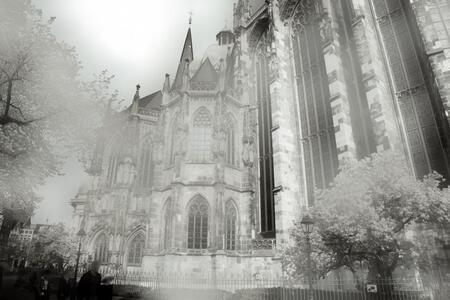}
\end{minipage}
\begin{minipage}{.13\textwidth}
    \centering
    \includegraphics[height=0.06\textheight]{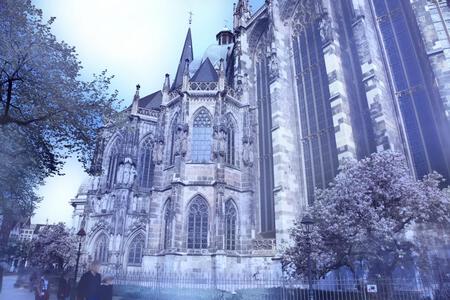}
\end{minipage}

\centering
\begin{minipage}{.13\textwidth}
    \centering
    \includegraphics[height=0.06\textheight]{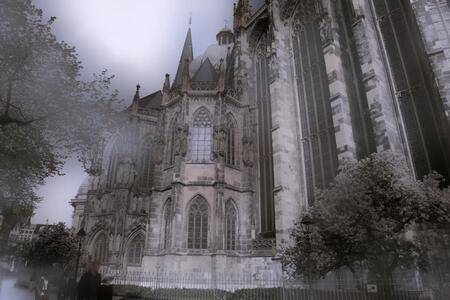}
\end{minipage}
\begin{minipage}{.13\textwidth}
    \centering
    \includegraphics[height=0.06\textheight]{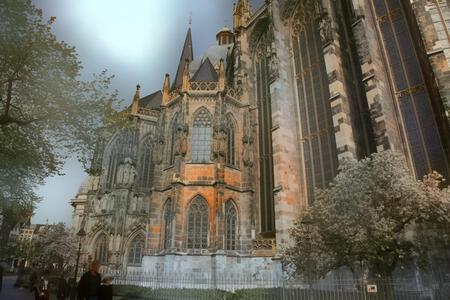}
\end{minipage}
\begin{minipage}{.13\textwidth}
    \centering
    \includegraphics[height=0.06\textheight]{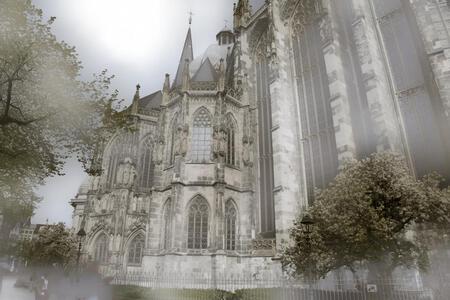}
\end{minipage}
\begin{minipage}{.13\textwidth}
    \centering
    \includegraphics[height=0.06\textheight]{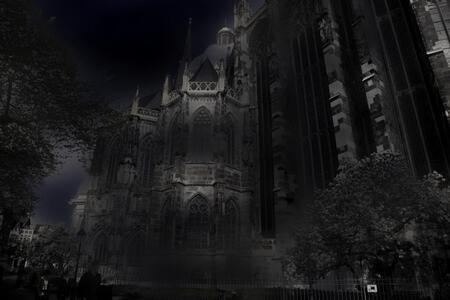}
\end{minipage}
\begin{minipage}{.13\textwidth}
    \centering
    \includegraphics[height=0.06\textheight]{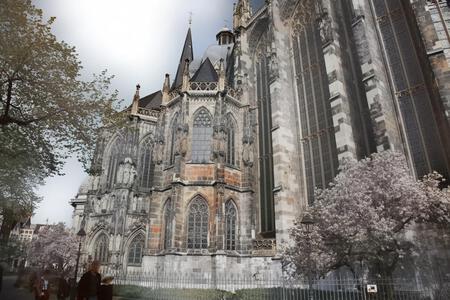}
\end{minipage}
\begin{minipage}{.13\textwidth}
    \centering
    \includegraphics[height=0.06\textheight]{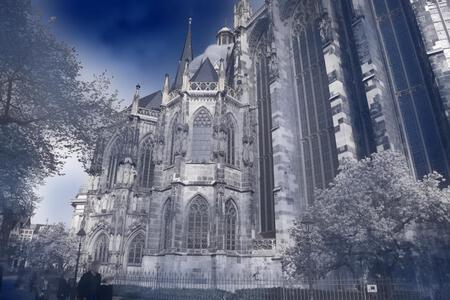}
\end{minipage}

\centering
\begin{minipage}{.13\textwidth}
    \centering
    \includegraphics[height=0.06\textheight]{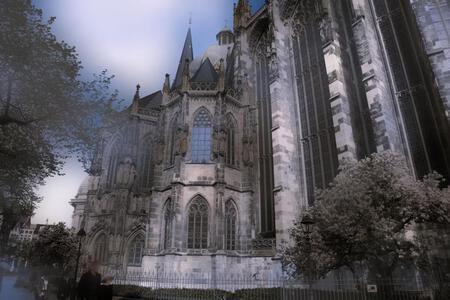}
\end{minipage}
\begin{minipage}{.13\textwidth}
    \centering
    \includegraphics[height=0.06\textheight]{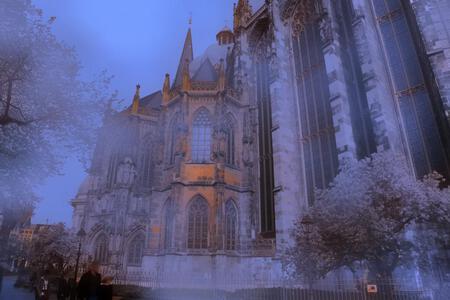}
\end{minipage}
\begin{minipage}{.13\textwidth}
    \centering
    \includegraphics[height=0.06\textheight]{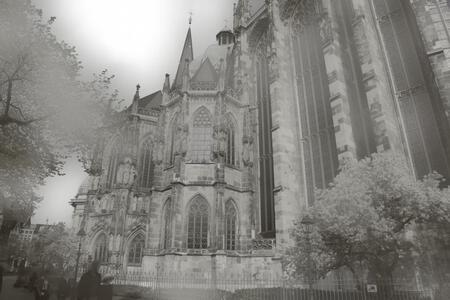}
\end{minipage}
\begin{minipage}{.13\textwidth}
    \centering
    \includegraphics[height=0.06\textheight]{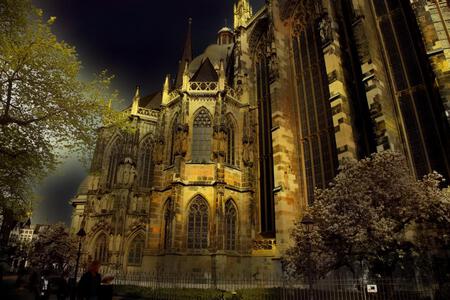}
\end{minipage}
\begin{minipage}{.13\textwidth}
    \centering
    \includegraphics[height=0.06\textheight]{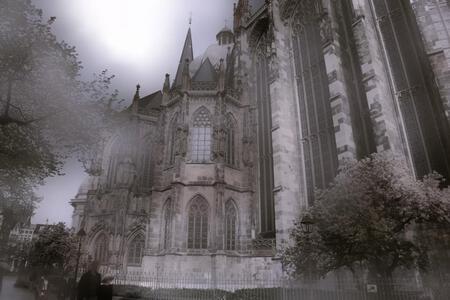}
\end{minipage}
\begin{minipage}{.13\textwidth}
    \centering
    \includegraphics[height=0.06\textheight]{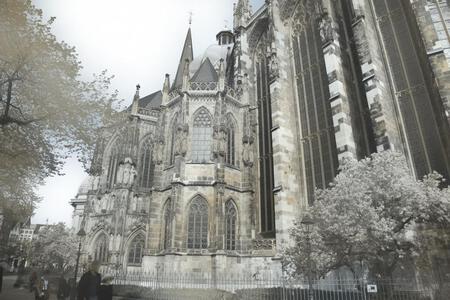}
\end{minipage}

\centering
\begin{minipage}{.13\textwidth}
    \centering
    cloudy
\end{minipage}
\begin{minipage}{.13\textwidth}
    \centering
    dusk
\end{minipage}
\begin{minipage}{.13\textwidth}
    \centering
    mist
\end{minipage}
\begin{minipage}{.13\textwidth}
    \centering
    night
\end{minipage}
\begin{minipage}{.13\textwidth}
    \centering
    rainy
\end{minipage}
\begin{minipage}{.13\textwidth}
    \centering
    snow
\end{minipage}

\caption{
From top to bottom:
(a) Day image from Aachen Day-Night benchmark.
(b) Night image stylization result used to train R2D2 model.
(c) Our image stylization results using style images from Figure~\ref{fig:styles}. Notice how different style images for the same category produce different results. Having multiple style images increases appearance variation of our training data.}\label{fig:stylized}
\end{figure*}

\subsection{Local Feature Model}
In this section we briefly discuss learned feature point method that we experimented with. The reader is encouraged to check original papers for R2D2~\cite{r2d2,r2d2_neurips}.

R2D2 architecture consists of L2-Net~\cite{l2net} backbone with two heads: last 128-dimensional feature layer divided by $L2$ norm to produce descriptors, and element-wise square of the last 128-dimensional feature layer followed by two separate 1x1 convolutions and softmax to produce a reliability mask and repeatability mask.

The neural network is trained on a batch of pairs of 192x192 crops of images, with known warp field between crops that are estimated either from a synthetic homography or 3D reprojection (computed by optical flow with epipolar constraint).

The objective is to optimize repeatability and reliability masks of feature points and the associated descriptors.

The repeatability loss encourages the repeatability mask to be similar for the pairs of crops, and encourages peakyness of masks:
\begin{equation}
\mathcal{L}_{rep}(I, I', U) = \mathcal{L}_{cosim}(I, I', U) + \lambda(\mathcal{L}_{peaky}(I) + \mathcal{L}_{peaky}(I'))
\end{equation}
where
both $\mathcal{L}_{rep}$ and $\mathcal{L}_{peaky}$ loss terms are enforced in a $N \times N$ window around each pixel. Hence, varying the value of $N$ changes the predicted repeatability mask, with smaller values of $N$ increasing the total number of detections in an image and larger values of N reducing the total number of detections and encouraging more repeatable points to be detected.

Instead of contrastive or triplet loss, R2D2 maximizes the average precision of descriptors in image $I$ that query image $I'$, so that the matching descriptor of image $I'$ is closest among all descriptors of $I'$ in terms of Euclidean distance. Hence, the reliability mask $\mathbf{R}$ and descriptors are optimized jointly as
\begin{equation}
\mathcal{L}_{AP_\kappa}(i, j) = 1 - \left[ AP(i,j) \mathbf{R}_{ij} + \kappa(1-\mathbf{R}_{ij})\right],
\end{equation}
where $\kappa=0.5$ is used to encourage the reliability mask to either pay a fixed penalty, or optimize the descriptors for better average precision. In our experiments we found that the training is more stable when we set $\kappa$ as the mean average precision of the batch if the mean average precision is lower than 0.5, \ie $\kappa=min(0.5, 0.5\sum_{i,j} AP(i,j))$.

As discussed in section~\ref{sec:intro}, internet images that were used to build a 3D model of a place may have varying appearance. In order to fairly measure the benefits of image stylization, we explicitly describe the training data.

At test time, we follow the procedure in R2D2. First, the image is repeatedly downsampled by a factor of $\sqrt[4]{2}$, until the maximum size of the image is 1024. Next, the image is repeatedly processed by the network and downsampled by a factor of $\sqrt[4]{2}$, until both dimensions of the image are less than 256. Non-maximum suppression is applied to the predicted repeatability mask at each scale. Pixel coordinates that have repeatability or reliability scores below 0.7 are removed. Preserved pixel locations across all scales are sorted by detection scores, which is a product of repeatability and reliability scores. A fixed number of top detections are extracted for feature matching.

\section{Experiments}
In this section we:
\begin{enumerate}
    \item Describe evaluation task used for measuring feature robustness,
    \item Investigate if image stylization provides better feature robustness than alternative approaches,
\end{enumerate}

\subsection{Evaluation Task}

Our experiments are focused on visual localization task.
We found that alternative benchmarks, such as homography estimation~\cite{hpatches_dataset} or Structure-from-Motion reconstructions~\cite{schonberger2017comparative} have limited appearance variation, which makes evaluation of robustness to appearance variations difficult.

Visual localization is the task of estimating a camera pose of a query image $Q$ in a pre-captured scene. A database of $N_{d}$ images with known poses is provided, and the algorithm is tested on a set of $N_{q}$ query images that do not overlap with the database images.

In this work, we focus on image-retrieval based methods, where for each query we first retrieve top $K=20$ images according to image-retrieval method such as DenseVLAD~\cite{vlad1,vlad2,densevlad} or NetVLAD~\cite{netvlad}, and then compute 2D-2D correspondences using detections and descriptions computed by local feature networks.
The pose of the query is solved using COLMAP's bundle adjuster using estimated correspondences and known poses of the database images as the constraint. The estimated pose is compared against ground-truth to compute success rate of localization. Localization attempt for a query is successful if the estimated camera position and camera orientation are both within certain thresholds. The localization success is measured at $(0.25m, 2^\circ)$, $(0.5m, 5^\circ)$, $(5m, 10^\circ)$ translation and rotation thresholds.

Feature matching is only one part of a visual localization system, other components such as image retrieval method can have large influence on the localization performance. Indeed, if image retrieval method is not robust to appearance changes, and retrieved images are not from the same part of the scene as the query image, then the localization attempt is bound for failure regardless of the features used to compute the pose. Hence, it is important to account for image retrieval method used for experiments. Unless otherwise specified, we use ``retrievalSfM120k-resnet101-gem'' model of Radenović~\etal~\cite{radenovic1,radenovic2}.

Furthermore, the number of features extracted from the query and/or the retrieved images also changes the localization performance. When low number of features is extracted from the images, then both the detector and descriptor need to be robust to appearance changes. However, if a large number of features is extracted, then the effect of the robustness of the detector is diminished, as the robustness of the descriptor becomes more important.

\paragraph{Aachen Day-Night v1.0~\cite{aachen1,sattler2018benchmarking}}
There are $N_{d}=4328$ database images captured in Aachen, Germany at daytime throughout two years with a hand-held camera. There are $N_{q}=824$ daytime query images and $N_{q}=98$ night-time query images captured with a mobile phone with software HDR.

The benchmark measures success for day queries and night queries separately. For night-time only competition, there is a list of image pairs that contain day-day and day-night image pairs, which one needs to compute correspondences for. However, if one is competing in day \emph{and} night tracks simultaneously, then the submission would also need to estimate top $K$ images for each query (if they the localization system is based on a image-retrieval method). Hence, the performance also depends on the method used to retrieve nearest images.

\paragraph{Aachen Day-Night v1.1~\cite{aachen11}}
This is an updated version of Aachen Day-Night challenge with a larger set of reference images and night-time queries. Specifically, there are $N_{d}=6697$ database images, $N_{q}=824$ daytime query images and $N_{q}=191$ night-time query images captured with a mobile phone.
Similarly to v1.0, the day and night queries are measured separately, and there is a separate list of image pairs provided by the benchmark for night-time only evaluation.

\paragraph{Extended CMU Seasons~\cite{cmuseasons,sattler2018benchmarking}}
Extended CMU Seasons dataset is built on CMU Visual Localization dataset~\cite{cmuseasons} and consists of $N_{d}=60937$ database images captured in Pittsburgh, US over a period of one year with cameras mounted on a car. There are $N_{q}=56613$ query images which consist of multiple sequences traverse by the car. Both database and query images depict urban, suburban and park scenes captured at different seasons.

\paragraph{SILDA Weather and Time of Day~\cite{silda}}
SILDA benchmark consists of $N_{d}=8334$ and $N_{q}=6064$ images captured on a spherical camera around Imperial College in London over a period of 12 months. The database consists of daytime images captured at clear and rainy weathers. The queries are made under 3 conditions: daytime images during snowy weather (labelled as snow), dusk-time images during clear days (evening) and night-time images during clear days (night).

\paragraph{RobotCar Seasons~\cite{robotcar,sattler2018benchmarking}}
RobotCar Seasons dataset is a subset of RobotCar dataset~\cite{robotcar} which consists of images captured by cameras mounted on a car, collected over a period of 12 months. There are 49 locations that do not overlap, and multiple capture conditions such as overcast, dawn, dusk, \etc for each location. The overcast footage with $N_{d}=26121$ frames is used as database images. The other conditions with and $N_{q}=11934$ frames are used as queries.

When localizing, we know each query's location (out of 49 locations) and so the image retrieval and pose estimation is done using database images corresponding to that location only.

\paragraph{RobotCar Seasons v2~\cite{robotcar,sattler2018benchmarking}}
RobotCar Seasons v2 dataset extends RobotCar Seasons v1 dataset~\cite{robotcar} by also releasing around half of query images from RobotCar Seasons v1 as database images. The rest of footage is used as query images.

\begin{table*}[ht]
\resizebox{\textwidth}{!}{ 
\begin{tabular}{l|cc|cc|ccc}
    & \multicolumn{2}{c|}{Aachen v1.1~\cite{aachen11}} & \multicolumn{2}{c|}{RobotCar~\cite{robotcar,sattler2018benchmarking}} & \multicolumn{3}{c}{SILDA~\cite{silda}}\\
    & day & night & day-overcast & other & evening & snow & night \\ \hline
5k R2D2-WA~\cite{r2d2} & $87.6 / 94.2 / 98.5$ & $61.8 / 79.1 / 91.6$ & $56.4 / 80.5 / 95.2$ & $12.8 / 26.3 / 36.0$ & $31.4 / 64.6 / 83.9$ & $0.2 / 14.6 / 63.2$ & $28.6 / 53.6 / 77.9$ \\
5k R2D2-WAS~\cite{r2d2} & $88.0 / 94.3 / 98.2$ & $66.0 / 80.1 / 96.3$ & not finished & not finished & $31.7 / 64.2 / 86.9$ & $2.8 / 14.2 / 66.3$ & $29.8 / 53.7 / 77.9$ \\
5k R2D2-WASF~\cite{r2d2} & $87.5 / 94.4 / 98.9$ & $70.2 / 84.3 / 96.3$ & $56.5 / 80.6 / 95.0$ & $19.1 / 34.3 / 46.1$ & $31.4 / 65.0 / 85.8$ & $1.4 / 15.4 / 63.9$ & $29.8 / 54.0 / 78.4$
 \\ \hline\hline
5k R2D2-WAS*F & $88.1 / 94.7 / 98.8$ & $68.1 / 84.3 / 96.9$ & $56.5 / 80.6 / 96.0$ & $20.6 / 40.2 / 52.4$ & $31.8 / 65.9 / 86.6$ & $2.6 / 14.6 / 64.2$ & $30.6 / 53.8 / 78.4$
 \\
5k R2D2-P & $87.9 / 94.7 / 98.5$ & $68.1 / 81.7 / 94.8$ & $56.3 / 80.7 / 95.9$ & $17.5 / 31.9 / 42.4$ & $31.6 / 65.2 / 85.1$ & $0.3 / 12.2 / 64.4$ & $28.7 / 53.4 / 78.4$  \\
5k R2D2-P-CA & $87.4 / 94.7 / 98.4$ & $66.5 / 83.8 / 95.8$ & $56.3 / 80.5 / 95.3$ & $20.8 / 38.4 / 50.2$ & $31.7 / 65.2 / 84.9$ & $1.0 / 12.7 / 64.4$ & $28.8 / 53.5 / 78.6$ \\
5k R2D2-P-O2S & $87.9 / 94.5 / 98.3$ & $72.3 / 88.0 / 97.4$ & $56.4 / 80.5 / 94.9$ & $21.6 / 40.2 / 53.0$ & $31.9 / 65.2 / 87.7$ & $2.9 / 14.9 / 67.8$ & $30.5 / 53.8 / 78.8$
\end{tabular}
}
\caption{Evaluation results for Aachen v.1.1, RobotCar and SILDa localization benchmarks. Scores represent localization success rate of queries for $(0.25m, 2^\circ)$, $(0.5m, 5^\circ)$, $(5m, 10^\circ)$ translation and rotation thresholds.}
\label{tab:loc_bench}
\end{table*}

\begin{table}[ht]
\centering
\resizebox{0.49\textwidth}{!}{ 
\begin{tabular}{l|cc|c}
    & \multicolumn{2}{c}{Aachen v1~\cite{aachen1}} 
    & \multicolumn{1}{c}{Aachen v1~\cite{aachen1}}
    \\
    & day & night & night-only \\ \hline
5k R2D2-WA~\cite{r2d2} & $84.3 / 92.5 / 96.8$ & $67.3 / 78.6 / 83.7$ & $61.2 / 74.5 / 84.7$ \\
5k R2D2-WAS~\cite{r2d2} & $85.2 / 93.4 / 97.0$ & $72.4 / 81.6 / 92.9$ & $69.4 / 83.7 / 95.9$ \\
5k R2D2-WASF~\cite{r2d2} & $85.3 / 93.1 / 97.3$ & $71.4 / 82.7 / 92.9$ & $65.3 / 81.6 / 94.9$ \\
\hline
5k R2D2-WAS*F & $85.9 / 93.7 / 97.5$ & $72.4 / 87.8 / 94.9$ & $73.5 / 85.7 / 100.0$ \\
5k R2D2-P & $85.0 / 93.1 / 97.1$ & $72.4 / 82.7 / 89.8$ & $66.3 / 76.5 / 84.7$ \\
5k R2D2-P-CA & $84.5 / 92.5 / 96.8$ & $75.5 / 83.7 / 91.8$ & $67.3 / 79.6 / 92.9$ \\
5k R2D2-P-O2S  & $84.7 / 92.8 / 96.8$ & $76.5 / 87.8 / 95.9$ & $73.5 / 85.7 / 99.0$
\end{tabular}
}
\caption{Evaluation results for Aachen Day-Night v.1 localization benchmark. We use image-retrieval based method to evaluate day and night queries. We also show evaluation with night queries only, using provided list of image pairs. Scores represent localization success rate of queries for $(0.25m, 2^\circ)$, $(0.5m, 5^\circ)$, $(5m, 10^\circ)$ translation and rotation thresholds.}
\label{tab:loc_bench2}
\end{table}

\subsection{Investigation of Appearance Augmentations}

In this section we compare image stylization to appearance augmentation strategies used in the literature: color augmentation~\cite{albumentations}, night style transfer in R2D2~\cite{r2d2} and multiple style transfers described in section~\ref{sec:method}. 

In this section compare trained models on Aachen v1.1, RobotCar Seasons v1 and SILDa benchmarks. For all 3 methods we use Radenovic~\etal~\cite{radenovic1,radenovic2} to extract top $K=20$ nearest neighbors for each query for feature matching. We extract 5000 keypoints from query and retrieved images in these experiments.

Next, we describe models that we used for comparison.

First, we used three baseline models of R2D2 (N=16).
R2D2 uses three types of images: random web images (W), day images from Aachen dataset (A), and night stylization to Aachen dataset day images (S). So,
\textbf{R2D2-WA} and \textbf{R2D2-WAS} are trained with synthetic homographies and color augmentations from corresponding image sets.
Finally, \textbf{R2D2-WASF} also uses optical flow between day images of Aachen dataset on top of \textbf{R2D2-WAS} training data. The total number of image pairs used to train \textbf{R2D2-WAS} and \textbf{R2D2-WASF} models is 3636.

R2D2-WA, R2D2-WAS and R2D2-WASF models were trained for 25 epochs with a fixed learning rate.

Next, we describe models trained by us:
\textbf{R2D2-WAS*F} is trained using 6 styles described in section~\ref{sec:method} to stylize Aachen day images to different style categories (see Figure~\ref{fig:stylized}), while web images only have color augmentation. The optical flow correspondences are used only between Aachen day and day images similar to R2D2-WASF. For each training epoch we randomly sample for the pool of stylized image so that the total number of batches in each epoch is the same as used to train R2D2-WASF model.

\textbf{R2D2-P}, \textbf{R2D2-P-CA}, and \textbf{R2D2-P-O2S}
models are trained on images from Phototourism dataset, where we randomly sample 300 images from each scene of the training set to produce a set of 3900 images.
\textbf{R2D2-P} is trained on these 3900 images and synthetic homographies without color augmentation. \textbf{R2D2-P-CA} is trained using synthetic homographies with color augmentation. For color augmentation we use Albumentations~\cite{albumentations} library to generate random color transforms of training images. \textbf{R2D2-P-O2S} is trained on images from Phototourism dataset and synthetic homographies with image stylization and color augmentation as described in section~\ref{sec:method}. For each training epoch we randomly sample one stylized image out of 60 available for each original image. So, each training epoch consists of 3900 pairs, which is only slightly more than 3636 used by baseline models.

R2D2-WAS*F, R2D2-P, R2D2-P-CA, and R2D2-P-O2S were trained for 70 epochs with exponentially decaying learning rate with warm-up of 5 epochs.

We evaluated these models on Aachen v1.1, RobotCar and SILDa benchmarks in table~\ref{tab:loc_bench}. We also show evaluation results for Aachen v1 in table~\ref{tab:loc_bench2}.

First, we can see that color augmentation is important, especially for RobotCar dataset, by comparing scores of R2D2-P and R2D2-P-CA.

If we compare R2D2-WA and R2D2-P-CA, we can see that Phototourism images and more aggressive color augmentation strategies help, especially for RobotCar and night queries of Aachen v1 and Aachen v1.1 benchmarks. One could hypothesize that Phototourism images are more varied than WA images, and may be some of the scenes are more similar to RobotCar images than W or A images. However, it is surprising that we can outperform R2D2-WA model with R2D2-P-CA model on Aachen challenges, especially when R2D2-WA has seen Aachen day images while R2D2-P-CA model has not seen them. This result suggests that tuning color augmentation during training is more important than images used for training.
There is, however, a slight score reduction of R2D2-P-CA compared to R2D2-WA for one of thresholds of snow queries of SILDa, but this is a very challenging category for all models.

It is clear that image stylization in addition to color augmentation is beneficial for localization when comparing R2D2-WA vs R2D2-WAS scores or R2D2-P-CA vs R2D2-P-O2S scores. The improvements are consistent and significant, especially for non-day queries in all benchmarks.

Next, we can see that stylizing images with multiple style categories and multiple style examples is very beneficial. Indeed, when comparing R2D2-WASF and R2D2-WAS*F the performance on Aachen v1.1 and SILDA datasets is comparable, but we get significant improvements in RobotCar dataset and night queries of Aachen v1 datasets.

Finally, we can see that when training with synthetic homographies and multiple style categories we can train R2D2-P-O2S model that is generally more competitive then R2D2-WASF model in the localization benchmarks. Note the difference in scores in Aachen v1, Aachen v1.1, RobotCar datasets.
This is significant in two ways. First, it means that we can train feature networks for localization task without SfM reconstructions. So, a feature network trained on raw images could potentially be used in an SfM pipeline to produce a 3D model with higher appearance coverage. Second, it suggests that for localization tasks, the 3D occlusion effects are not as significant as appearance variations.

\section{Challenges}
\subsection{Long-Term Visual Localization at ECCV 2020}
The challenge focuses on visual localization in 3 conditions/tracks:
\begin{enumerate}
    \item Visual Localization for Autonomous Vehicles: where localization systems submitted to the competition are evaluated on Extended CMU Seasons, RobotCar Seasons v2, and SILDa Weather and Time of Day datasets. 
    \item Visual Localization for Handheld Devices: Aachen Day-Night v1.1 and InLoc~\cite{inloc,inloc2} datasets are used for evaluation of submissions.
    \item Local Features for Long-term Localization: the submissions are evaluated using night-time queries of Aachen Day-Night v1.1 benchmark using provided image pairs.
\end{enumerate}
In each track, the submissions are evaluated on listed datasets for all 3 thresholds. The ranking of localization systems is determined with Schulze method.

Our submission uses R2D2 features trained on images from Phototourism dataset~\cite{imw_challenge} using synthetic homographies and image stylization (original to stylized).
Except for ``Local Features for Long-term Localization'' track which provides image pairs, we use image-retrieval based localization method. So, for each query image, top $K=20$ images are retrieved from the database using Radenovic~\etal~\cite{radenovic1,radenovic2}.

We extract 10000 features from images for ``Local Features for Long-term Localization'' challenge and 5000 for other challenges (both query and retrieved images) . See tables~\ref{tab:ltvl:auto},~\ref{tab:ltvl:hand}, and~\ref{tab:ltvl:local} for scores in each of the tracks, respectively.

We note that we only used a single query image for localization in ``Autonomous Vehicles'' track, while the benchmark allows to localize from multiple frames. For example, in SILDa dataset the query image is one of the two spherical images of the camera rig, which means that the camera pose of the query image can be estimated by querying the database using opposite view of the camera rig.

\begin{table*}[ht]
\resizebox{\textwidth}{!}{ 
\begin{tabular}{l|ccc|cc|ccc}
    & \multicolumn{3}{c|}{Extended CMU Seasons~\cite{cmuseasons,sattler2018benchmarking}} & \multicolumn{2}{c|}{RobotCar Seasons v2~\cite{robotcar,sattler2018benchmarking}} & \multicolumn{3}{c}{SILDA~\cite{silda}}\\
    & urban & suburban & park & day all & night all & evening & snow & night \\ \hline
5k R2D2-P-O2S & $93.8 / 96.6 / 98.2$ & $83.5 / 86.8 / 90.5$ & $76.4 / 80.9 / 85.7$ & $0.0 / 0.0 / 0.0$ & $0.0 / 0.0 / 0.0$ & $31.9 / 65.2 / 87.7$ & $2.9 / 14.9 / 67.8$ & $30.5 / 53.8 / 78.8$
\\
\end{tabular}
}
\caption{Scores in ECCV 2020 Long-Term Visual Localization for Autonomous Vehicles challenge.}
\label{tab:ltvl:auto}
\end{table*}

\begin{table*}[ht]
\centering
\resizebox{0.7\textwidth}{!}{ 
\begin{tabular}{l|cc|cc}
    & \multicolumn{2}{c|}{Aachen v1.1~\cite{aachen11}} & \multicolumn{2}{c}{InLoc~\cite{inloc,inloc2}} \\
    & day & night & duc1 & duc2 \\ \hline
5k R2D2-P-O2S & $87.1 / 94.7 / 98.3$ & $74.3 / 86.9 / 97.4$ & $39.4 / 58.1 / 70.2$ & $41.2 / 61.1 / 69.5$
\\
\end{tabular}
}
\caption{Scores in ECCV 2020 Long-Term Visual Localization for Handheld Devices challenge.}
\label{tab:ltvl:hand}
\end{table*}

\begin{table}[ht]
\centering
\begin{tabular}{l|c}
    & \multicolumn{1}{c}{Aachen v1.1~\cite{aachen11}} \\
    & night \\ \hline
10k R2D2-P-O2S & $69.6 / 86.9 / 97.9$ \\
\end{tabular}
\caption{Scores in ECCV 2020 Long-Term Visual Localization Local feature challenge.}
\label{tab:ltvl:local}
\end{table}

\subsection{Map-based Localization for Autonomous Driving}
This challenge also evaluates a localization system, where the task is to estimate 6 degrees of freedom relative pose between queries and database images. The images are collected using car-mounted grayscale stereo camera.

\begin{figure*}[ht]
\centering
\begin{minipage}{.3\textwidth}
    \centering
    \includegraphics[width=0.99\textwidth]{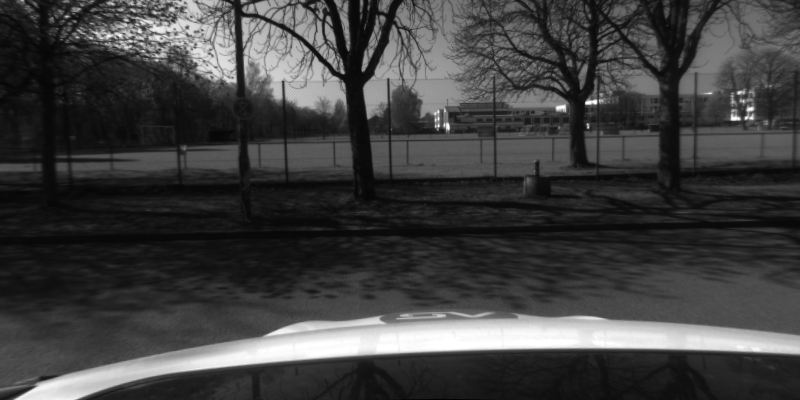}
\end{minipage}
\begin{minipage}{.3\textwidth}
    \centering
    \includegraphics[width=0.99\textwidth]{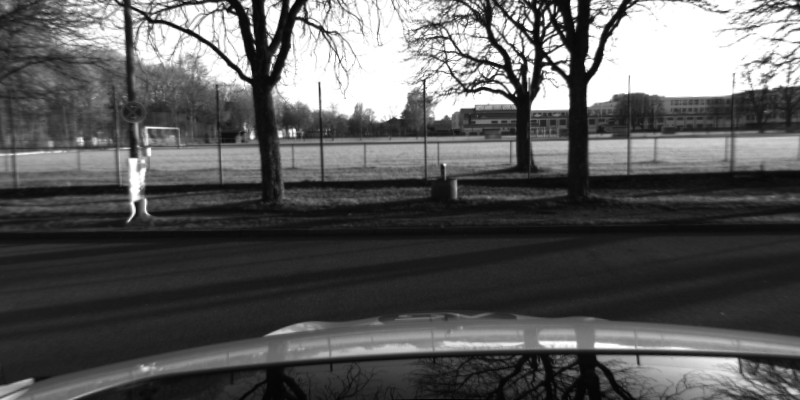}
\end{minipage}
\begin{minipage}{.3\textwidth}
    \centering
    \includegraphics[width=0.99\textwidth]{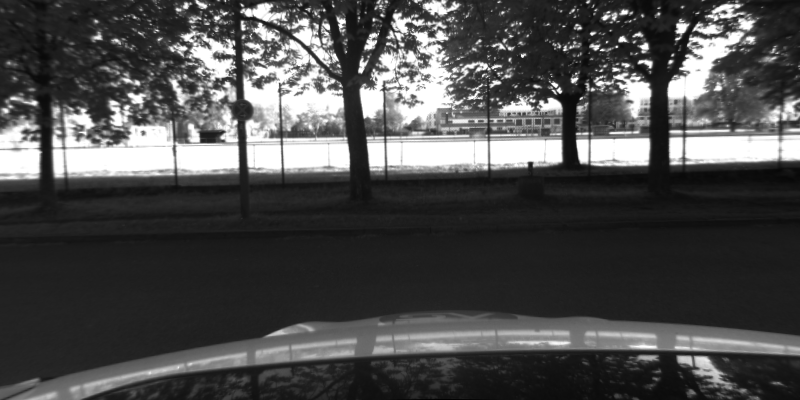}
\end{minipage}

\centering
\begin{minipage}{.3\textwidth}
    \centering
    \includegraphics[width=0.99\textwidth]{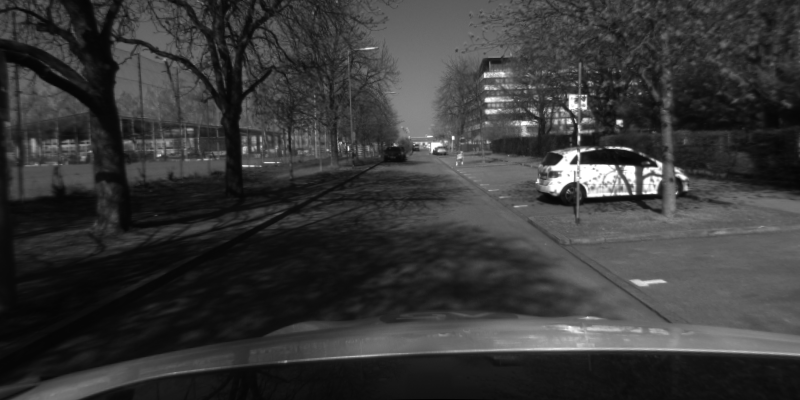}
\end{minipage}
\begin{minipage}{.3\textwidth}
    \centering
    \includegraphics[width=0.99\textwidth]{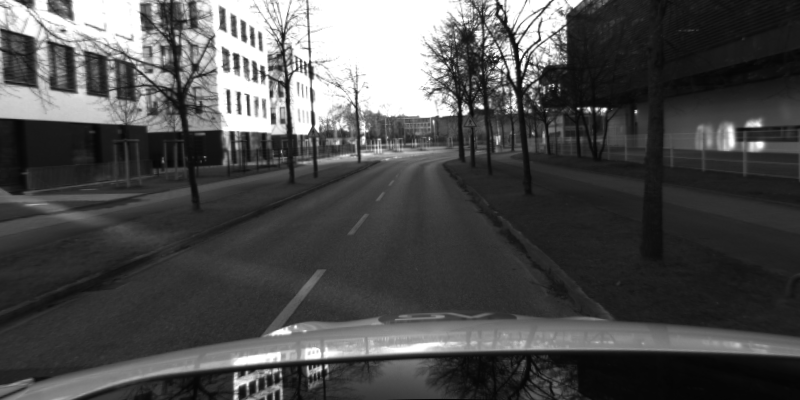}
\end{minipage}
\begin{minipage}{.3\textwidth}
    \centering
    \includegraphics[width=0.99\textwidth]{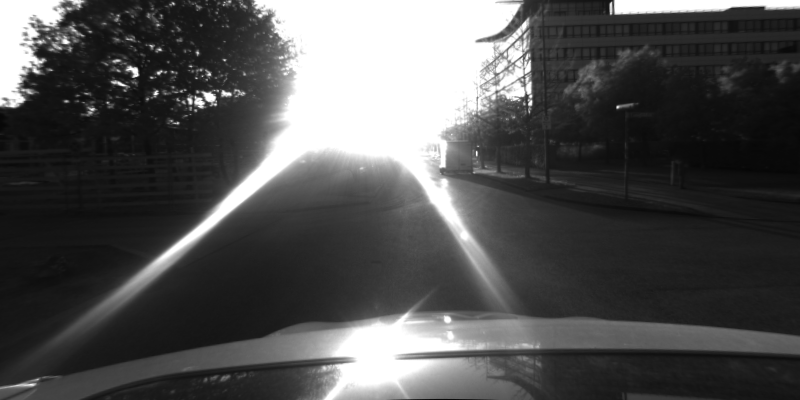}
\end{minipage}

\centering
\begin{minipage}{.3\textwidth}
    \centering
    \includegraphics[width=0.99\textwidth]{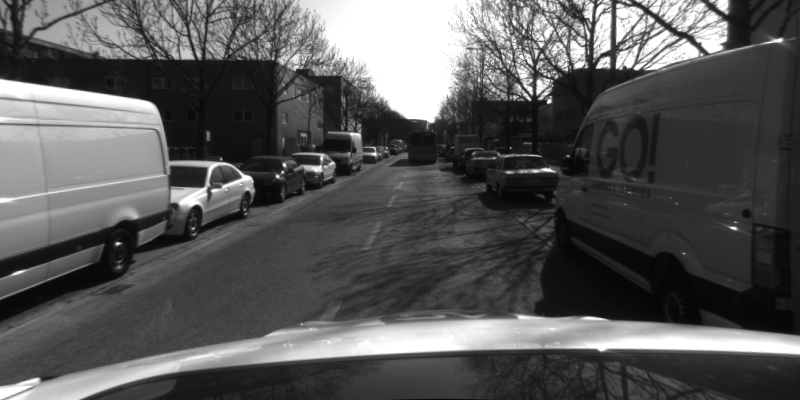}
\end{minipage}
\begin{minipage}{.3\textwidth}
    \centering
    \includegraphics[width=0.99\textwidth]{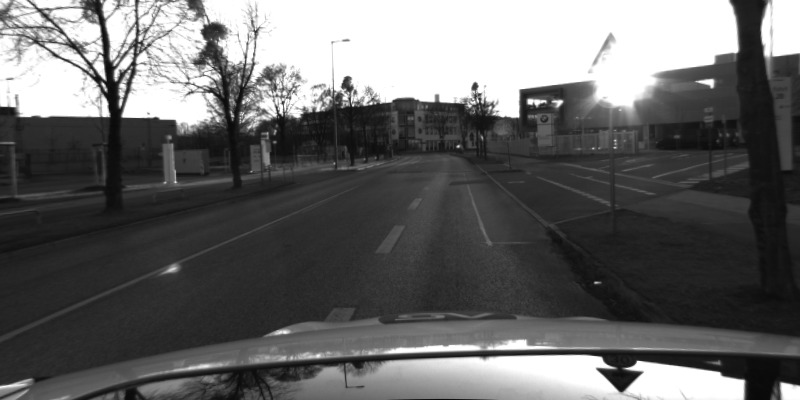}
\end{minipage}
\begin{minipage}{.3\textwidth}
    \centering
    \includegraphics[width=0.99\textwidth]{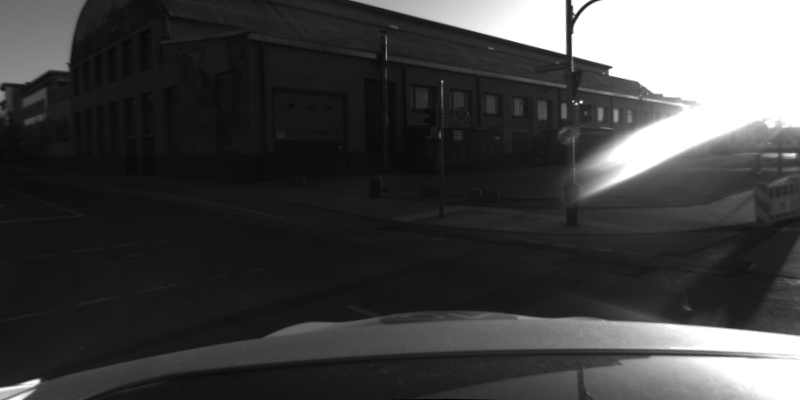}
\end{minipage}

\centering
\begin{minipage}{.3\textwidth}
    \centering
    \includegraphics[width=0.99\textwidth]{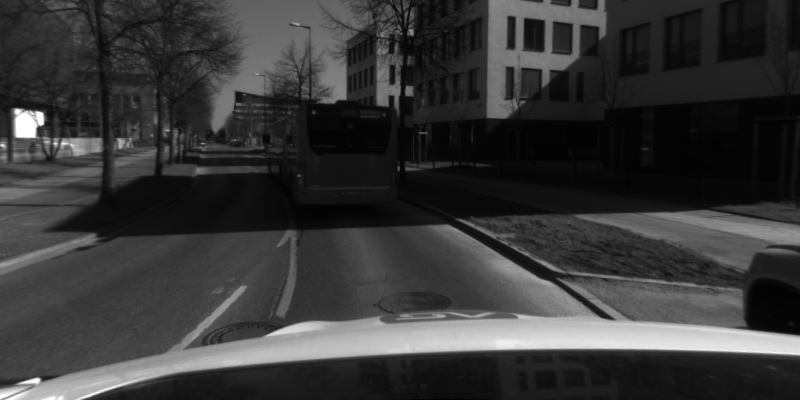}
\end{minipage}
\begin{minipage}{.3\textwidth}
    \centering
    \includegraphics[width=0.99\textwidth]{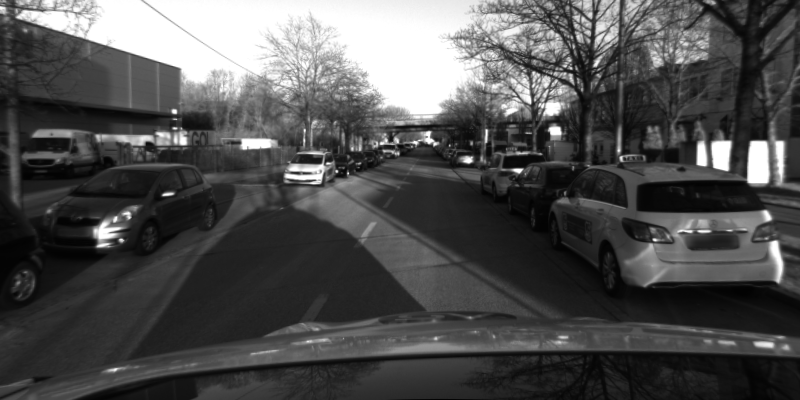}
\end{minipage}
\begin{minipage}{.3\textwidth}
    \centering
    \includegraphics[width=0.99\textwidth]{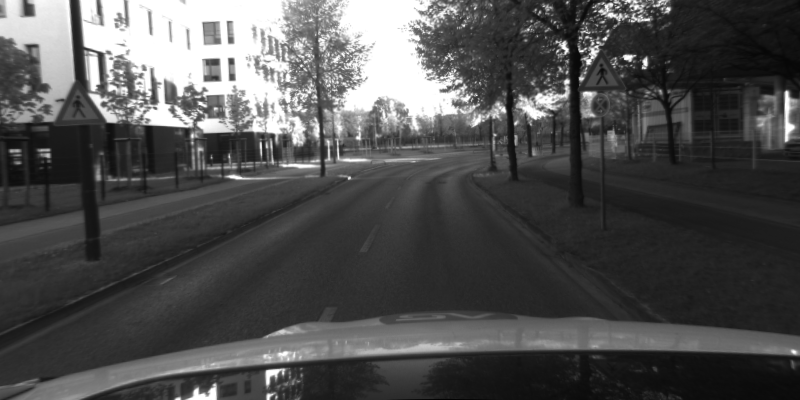}
\end{minipage}

\vspace{2pt}
\centering
\begin{minipage}{.3\textwidth}
    \centering
    (a) Reference
\end{minipage}
\begin{minipage}{.3\textwidth}
    \centering
    (b) test\_sequence0
\end{minipage}
\begin{minipage}{.3\textwidth}
    \centering
    (c) test\_sequence1
\end{minipage}

\caption{Random sample of reference and query images of MLAD challenge. The images show significant illumination variations, hard shadows, specularities and lense flares.}
\label{fig:mlad}
\end{figure*}

The database images are stereo video frames from one continuous sequence captured on the 7-th of April 2020 around 10:20. The query images are frames from two sequences captured on the 24-th of March around 17:45 and 23-rd of April around 19:37. Hence, robustness to illumination change and weather change is important for this challenge. Please see Figure~\ref{fig:mlad} for examples.

The evaluation measures localization success rate, \ie the percentage of queries where the translation error between the estimated and ground-truth poses is within a threshold. The accuracy is measured at 0.1m, 0.2m, and 0.5m thresholds.

The list of image pairs for estimating correspondences is provided by the benchmark.

Our scores together with baseline scores of competing method can be seen in table~\ref{tab:mlad}. Notice how additional stylization augmentation during training noticeably improves scores for `test\_sequence1', which is more challenging than `test\_sequence0'.

\begin{table*}[ht]
\centering
\begin{tabular}{l|c|c}
    & test\_sequence0 & test\_sequence1 \\
    \hline
Superpoint & $15.5 / 27.5 / 47.5$ & $9.0 / 19.4 / 36.4$ \\
Superpoint + SuperGlue & $21.2 / 33.9 / 60.0$ & $12.4 / 26.5 / 54.4$ \\
D2-Net & $12.5 / 29.3 / 56.7$ & $7.5 / 21.4 / 47.7$ \\
5k R2D2-WASF N=16 & $21.5 / 33.1 / 53.0$ & $12.3 / 23.7 / 42.0$ \\
\hline
5k R2D2-P-O2S N=16 & $20.9 / 33.2 / 54.8$ & $13.2 / 25.3 / 45.7$ \\
10k R2D2-P-O2S N=8 & $22.3 / 34.1 / 57.4$ & $13.2 / 26.0 / 47.8$ \\
\end{tabular}
\caption{Scores in Map-based Localization for Autonomous Driving challenge.}
\label{tab:mlad}
\end{table*}

\section{Conclusion}
In this work we investigate if image stylization can improve robustness of local features to illumination, weather and season changes. We experimented with multiple appearance augmentation techniques and demonstrated better robustness when augmentation with image stylization is used during training. We also show that stylizing images with multiple styles and style examples is better than stylizing with a few examples. Our models are competitive in visual localization benchmarks, outperforming baseline models, despite training without 3D correspondences.

Our experiments were done with R2D2 model, but in future versions of this paper we plan to extended experiments by evaluate other local feature networks, such as D2-Net~\cite{d2net} and Superpoint~\cite{superpoint}, when trained with stylized images. %

{\small
\bibliographystyle{ieee_fullname}
\bibliography{egbib}
}

\end{document}